\documentclass[10pt,twocolumn,letterpaper]{article}

\usepackage[pagenumbers]{cvpr}

\usepackage{graphicx}
\usepackage{amsmath}
\usepackage{amssymb}
\usepackage{booktabs}
\usepackage[normalem]{ulem}
\useunder{\uline}{\ul}{}
\usepackage{caption}
\usepackage{subcaption}

\usepackage[pagebackref,breaklinks,colorlinks]{hyperref}

\usepackage[capitalize]{cleveref}
\crefname{section}{Sec.}{Secs.}
\Crefname{section}{Section}{Sections}
\Crefname{table}{Table}{Tables}
\crefname{table}{Tab.}{Tabs.}

\def\cvprPaperID{917}
\def\confName{CVPR}
\def\confYear{2023}

\begin{document}

\title{Delving into Transformer for
Incremental Semantic Segmentation}

\author{Zekai Xu$^{1,2}$, Mingyi Zhang$^{4}$, Jiayue Hou$^{2}$, Xing Gong$^{2}$, Chuan Wen$^{3}$, Chengjie Wang$^{2}$, Junge Zhang$^{4}$ \\
$^{1}$ Department of Computer Science and Technology, Tsinghua University \\
$^{2}$ YouTu Lab, Tencent \\ 
$^{3}$ Institute for Interdisciplinary Information Sciences, Tsinghua University \\
$^{4}$ Institute of Automation, Chinese Academy of Sciences \\
{\tt\small \{xuzk20,cwen20\}@mails.tsinghua.edu.cn,
zhangmingyi2014@ia.ac.cn, }\\
{\tt\small \{jiayuehou,xaviergong,jasoncjwang\}@tencent.com,
jgzhang@nlpr.ia.ac.cn}
}
\maketitle

\begin{abstract}
   Incremental semantic segmentation(ISS) is an emerging task where old model is updated by incrementally adding new classes. At present, methods based on convolutional neural networks are dominant in ISS. However, studies have shown that such methods have difficulty in learning new tasks while maintaining good performance on old ones (\textsl{catastrophic forgetting}). In contrast, a Transformer based method has a natural advantage in curbing catastrophic forgetting due to its ability to model both long-term and short-term tasks. In this work, we explore the reasons why Transformer based architecture are more suitable for ISS, and accordingly propose propose \textsl{TISS}, a \textbf{T}ransformer based method for \textbf{I}ncremental \textbf{S}emantic \textbf{S}egmentation. In addition, to better alleviate catastrophic forgetting while preserving transferability on ISS, we introduce two patch-wise contrastive losses to imitate similar features and enhance feature diversity respectively, which can further improve the performance of \textsl{TISS}. Under extensive experimental settings with Pascal-VOC 2012 and ADE20K datasets, our method significantly outperforms state-of-the-art incremental semantic segmentation methods.
\end{abstract}

\section{Introduction}
\label{sec:intro}
Semantic segmentation is a fundamental problem in computer vision with a wide range of applications covering robotics, autonomous driving, medical image processing, etc. Over the past few years, semantic segmentation models based on Fully Convolutional Networks(FCNs) \cite{long2015fully} which extend deep architectures from image-level to pixel-level classification have made great progress \cite{chen2018encoder,lin2017refinenet,long2015fully,zhang2018exfuse,zhao2017pyramid}. However, current Convolutional Neural Networks(CNNs) based semantic segmentation approaches are typically based on training in a single-shot requiring whole datasets with the complete annotations on all seen categories. This issue is well-known for CNNs and it is called \textsl{catastrohpic forgetting}\cite{french1999catastrophic,goodfellow2013empirical,mccloskey1989catastrophic}, as CNNs based architecture struggles in incrementally updating their parameters for learning new classes whilest preserving the good performance on the old ones. This setup, referred here as Incremental Semantic Segmentation(denoted as ISS), has emerged very recently for specialized applications before being proposed for general segmentation datasets. 

Incremental learning has been widely studied in image classification \cite{kirkpatrick2017overcoming,li2017learning} and object detection\cite{li2019rilod,shmelkov2017incremental}, while has been tackled only recently for semantic segmentation\cite{cermelli2020modeling,douillard2021tackling,michieli2019incremental,michieli2021continual,tasar2019incremental}. However, current incremental semantic segmentation methods are commonly based on CNNs models. With the rapid development of Vision Transformer(ViT) \cite{dosovitskiy2020image,touvron2021training} on semantic segmentation\cite{bao2021beit,strudel2021segmenter}, we conduct an investigation on ISS and cope with existing problems by introducing \textsl{TISS}(A \textbf{T}ransformer based method for \textbf{I}ncremental \textbf{S}emantic \textbf{S}egmentation) through following steps. 

First, it is generally accepted that there are two reasons for the severe catastrophic forgetting of CNNs based architecture for ISS: (\romannumeral1) CNNs based architecture struggles in adapting data distribution shift between two tasks due to extensive application of Batch Normalization(BN)\cite{ioffe2015batch} layers, (\romannumeral2) the local nature of the convolution filter constrains the extraction of global information which is particularly important on semantic segmentation\cite{strudel2021segmenter}. However, Transformer\cite{vaswani2017attention} based architecture circumvents the limitations of CNNs based architecture as follows: (\romannumeral1) application of Layer Normalization(LN)\cite{ba2016layer} layers rather than BN layers avoids the impact of variability in data distribution across different tasks, (\romannumeral2) the self-attention layer in ViT provides better inductive bias to capture the global information than the convolution operation in CNN\cite{raghu2021vision}. Second, we innovatively combine the distillation based method with a Transformer architecture for ISS and achieve competitive performance. Third, to better alleviate the catastrophic forgetting, we introduce a patch-wise contrastive distillation loss $\mathcal{L}_\text{cd}$ specially for Transformer based architecture through consolidating the knowledge accumulation between the patch representations extracted from the current model and the previous model. Furthermore, we propose a patch-wise contrastive loss $\mathcal{L}_\text{ct}$ specially for the current model to enhance the generalization ability on new tasks. We extensively evaluate our method on two datasets, Pascal-VOC\cite{pascal-voc-2012} and ADE20K\cite{zhou2017scene}, showing that our method largely surpasses current state-of-the-art method for ISS.

To summarize, our contributions are three-fold:
\begin{itemize}
    \item[$\bullet$] We introduce a Transformer based method for ISS and analyse the advantages of Transformer based architecture on ISS. We innovatively combine the distillation based method with a Transformer architecture to cope with incremental learning problems on semantic segmentation.
    \item[$\bullet$] To further improve the performance of Transformer based methods in semantic segmentation tasks, we introduce two novel patch-wise contrastive losses, which can improve model expression ability and alleviate catastrophic forgetting by imitating similar features and enhancing feature diversity.
    \item[$\bullet$] We benchmark our method over several previous methods on Pascal-VOC and ADE20K, which leads to state-of-the-art considering different experimental settings. We hope that our results will benefit future works.
\end{itemize}

\section{Related work}

\subsection{Incremental Semantic Segmentation(ISS)}

In recent years, semantic segmentation has achieved outstanding results with deep CNNs and transformer methods. These approches have exploited multiple feature extraction and context priors strategies. However, they are not able to continuously learn new classes without retraining from scratch and experience catastrophic forgetting which is quite limited in practical applications. To solve this problem, incremental learning in semantic segmentation has been studied in past few years, especially Class-Incremental Learning(Class-IL). 

Recently, several works\cite{cermelli2021incremental,douillard2021tackling,klingner2020class,li2017learning,maracani2021recall,michieli2019incremental,michieli2021continual,michieli2021knowledge,qu2021recent,yan2021framework,yang2022continual} focused on incremental learning on semantic segmentation. \textsl{LwF}\cite{li2017learning} proposed a knowledge distillation method joint training with finetuning. Michieli et al.\cite{michieli2019incremental}presented some knowledge distillation based approaches on the previous models, while updating the new model to learn new ones with designing output-level and features-level distillation losses. \textsl{SDR}\cite{michieli2021continual} proposed several strategies like prototype matching, contrastive learning in the latent space to improve knowledge distillation. \textsl{RECALL}\cite{maracani2021recall},instead, proposed a replay method using samples of old categories to reduce the forgetting. \textsl{PLOP}\cite{douillard2021tackling} and \textsl{MiB}\cite{cermelli2020modeling} tackled the catastrophic forgetting by handling the background semantic distribution shift problem. \textsl{PLOP}\cite{douillard2021tackling} proposed to preserve long and short-range spatial relationships at feature level and generate pseudo-labels of background from previous model to mitigate forgetting. \textsl{MiB}\cite{cermelli2020modeling} proposed unbiased cross-entropy and output-level knowledge distillation losses with an unbiased decoder parameter initialization strategy, which is used in our method.

\subsection{Vision Transformer on semantic segmentation}
\label{sec: vit}

After being the state of the art in Natural Language Processing(NLP), transformer has gained rapid development and showed potential power in computer vision tasks. The Vision Transformer(ViT)\cite{dosovitskiy2020image} first introduced a transformer architecture without convolution for image classification and input images were treated as a sequence of tokens to multiple layers. More recent approaches such as \textsl{DeiT}\cite{touvron2021training}, \textsl{CPVT}\cite{chu2021conditional}, \textsl{TNT}\cite{han2021transformer} proposed various improved strategies based on ViT. In semantic segmentaion, \textsl{SegFormer}\cite{xie2021segformer} designed a network with encoder using a hierarchical pyramid vision transformer to obtain semantic segmentation mask. \textsl{Segmenter}\cite{strudel2021segmenter} proposed a mask transformer to enhance decoding performance with learnable class-map tokens. \textsl{MaskFormer}\cite{cheng2021per} and \textsl{Mask2Former}\cite{cheng2021masked} created an all-in-one mudule dealing with per-pixel classification in semantic segmentation. \textsl{SeMask}\cite{jain2021semask} proposed a framework to improve the finetuning ability of the pretrained vision transformer backbone and achieved the state of the art on ADE20k.

Recently, to further improve performance of transformer on computer vision tasks, some works focused on exploiting feature diversity of transformer. \cite{touvron2021going} found the training instability of transformer especially with wider and deeper model. \cite{gong2021vision} further studied the phenomenon in the perspective of patch diversification. They found that the patch-wise cosine similarity of patch representations for high layers in \textsl{DeiT}\cite{touvron2021training} and \textsl{Swin-Tansformer}\cite{liu2021swin} increased heavily which reduced the learning capacity of transformer. To enhance the learning capacity of transformer, they proposed three diversity-encouraging techniques to improve the capacity of transformer. The similar patch-wise auxiliary loss introduced in \cite{jiang2021token} also shed light on the importance and necessities of the feature diversity. In this work, we propose \textsl{TISS}, a Transformer based framework for ISS with two patch-wise contrastive losses tailored to it. Both of the transformer backbone and losses are effective in suppressing catastrohopic forgetting while enhancing the performance on new adding tasks, which improve overall performance significantly.

\section{Our method: \textsl{TISS}}

\begin{figure*}[h]
\centering
\includegraphics[width=\linewidth]{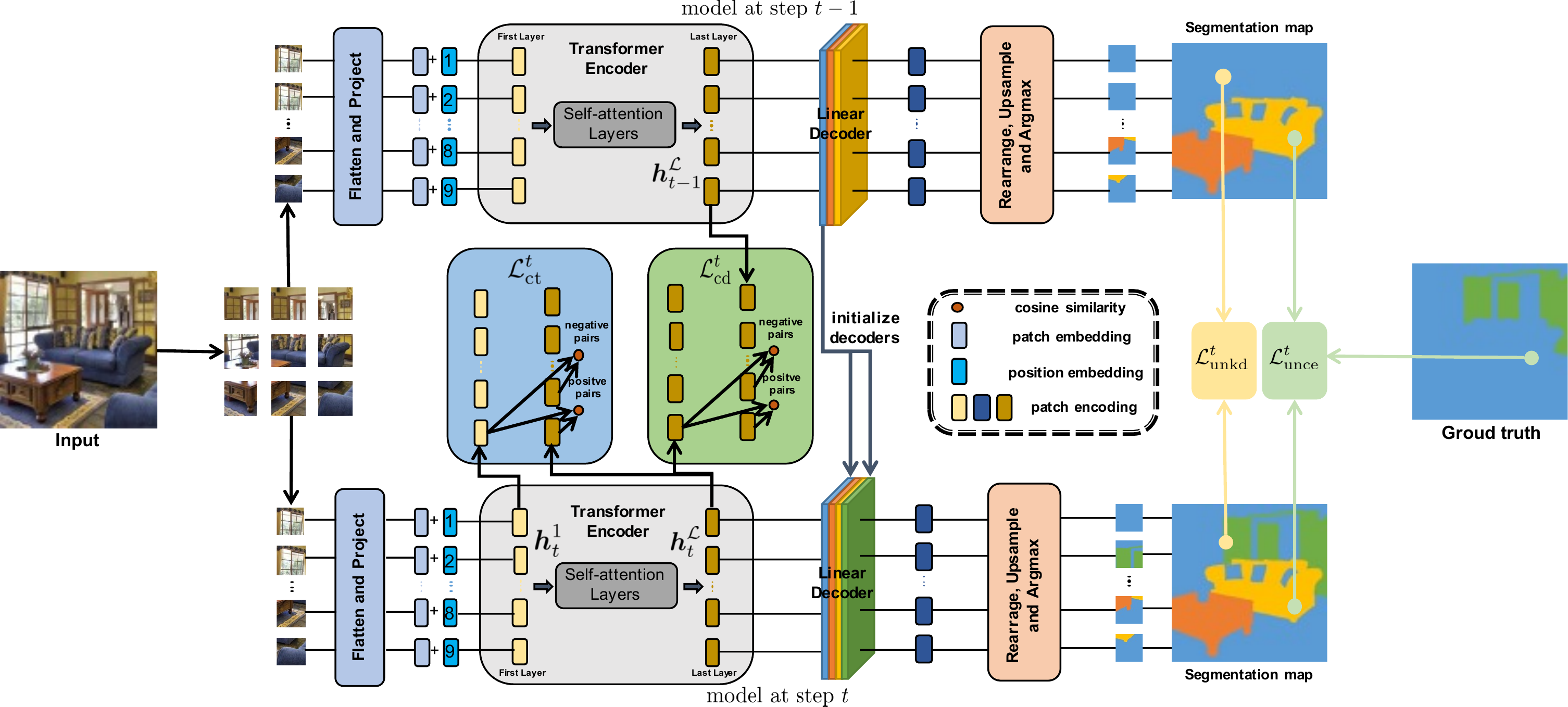}
\vskip -0.35cm
\caption{Overview of our method \textbf{\textsl{TISS}}. At learning step $t$, input image is processed by old(top) and current(bottom) models, generating the segmentation maps respectively. Our contributions are as follows: (\romannumeral1) combining the distillation based method \textbf{\textsl{MiB}} with a Transformer architecture, following the initialization strategy(blue lines), the unbiased knowledge distillation loss $\mathcal{L}^t_\text{unkd}$(light yellow block) and the unbiased cross entropy loss $\mathcal{L}^t_\text{unce}$(light green block). (\romannumeral2) Proposing a patch-wise contrastive distillation loss $\mathcal{L}^t_\text{cd}$(green block) tailored to Transformer based architecture to better alleviate catastrophic forgetting problems on ISS. (\romannumeral3) Introducing a patch-wise contrastive loss $\mathcal{L}^t_\text{ct}$(blue block) specially for Transformer to enhance the performance on the the new task by promoting patch diversification. } 
\label{fig:framework}
\vskip -0.5cm
\end{figure*}

\textsl{TISS} is a \textbf{T}ransformer based method for \textbf{ISS}. An overview of the method is shown in Figure \ref{fig:framework}. In this section we start by the problem definition of ISS. Then we derive the advantages of Transformer based architecture on suppressing catastrophic forgetting in ISS. Furthermore, we combine the distillation based method \textsl{MiB} with a Transformer architecture for ISS to alleviate forgetting problems. Finally, we put forward two novel patch-wise contrastive losses specifically for Transformer to further suppress catastrophic forgetting and enhance the performance on new adding tasks, respectively.

\subsection{Problem Definition}
To facilitate understanding, here we first introduce the definition of semantic segmentation. In semantic segmentation, we get a training set $\mathcal{T}\subset \mathcal{X} \times {\mathcal{Y}}$, where $\mathcal{X} \subset \mathbb{R} ^{H\times W\times3}$ denoted as input space, $\mathcal{Y} \subset \mathcal{C} ^{H\times W}$ denoted as output space, $\mathcal{C}$ as a collection of possible semantic classes. Given an image $x \in \mathcal{X}$ composed by a set of pixels $\mathcal{I}$ with constant cardinality $|\mathcal{I}|=N$, we aim at assigning each pixel ${x_i}$ a semantic class $\mathcal{C}_i \in \mathcal{C}$ or a special background category denoted as $b \in \mathcal{C}$. The mapping is realized with a designed parameterized model $\mathcal{M}:\mathcal{X}\mapsto \mathbb{R} ^{H\times W\times\mathcal{C}}$ learned in one shot and training set available at once. The output segmentation mask is obtained as $y^{*}=\left\{\arg \max _{c \in \mathcal{Y}} \mathcal{M}(x)[i, c]\right\}_{i=1}^{N}$, where $\mathcal{M}(x)[i, c]$ is the probability for class $c$ in pixel $i$.

In ISS, instead, the training is performed in multiple steps, denoted as $t=1...T$. At the $t_{th}$ learning step, only a subset of training data $\mathcal{T}_t\subset \mathcal{X} \times {\mathcal{Y}_t}$ is available, where $\mathcal{Y}_t \subset {\mathcal{S}_t}^{H\times W}$, $\mathcal{S}_t$ is the new classes set. Then an updated model $\mathcal{M}_{t}:\mathcal{X}\mapsto \mathbb{R} ^{H\times W\times\mathcal{C}_t}$ is learnt using the training subset $\mathcal{T}_t$ in conjunction with the previous model $\mathcal{M}_{t-1}:\mathcal{X}\mapsto \mathbb{R} ^{H\times W\times\mathcal{C}_{t-1}}$, $\mathcal{C}_t=\mathcal{C}_{t-1}\cup\mathcal{S}_t$, $\mathcal{C}_1=\mathcal{S}_1$. As in the standard incremental learning, we assume the sets of new categaries at each step to be disjoint with exception of the special background class $b$. 

\subsection{Why choose Transformer?}

Empirically, CNNs based architecture is commonly used as feature extractors in computer vision. However, considering ISS, the local nature of CNNs based architecture limits its performance as follows: (\romannumeral1) the extensive application of Batch Normalization(BN) layers in CNNs leads to catastrophic forgetting. BN layers are affected by data distribution during training, while the distribution of datasets for old and new tasks in incremental learning are often discrepant. It has been observed that the catastrophic forgetting problem is significantly alleviated when all Batch Normalization(BN) layers are fixed during training\cite{li2022technical,michieli2019incremental}. (\romannumeral2) The local nature of the convolution filter also limits the extraction of global information in the image which is particularly important for segmentation where the labelling of local patches often depends on the global image context\cite{strudel2021segmenter}. Meanwhile, limitations of global information extraction on the previous tasks would make the catastrophic forgetting on ISS even more difficult to circumvent, leading to further degradation.

The Transformer based architecture, however, is capable to circumvent the limitations of CNNs based architecture on ISS as follows: (\romannumeral1) the extensive application of Layer Normalization(LN) layers rather than Batch Normalization(BN) layers avoids the impact of variability in data distribution across different tasks on model training. (\romannumeral2) Compared to CNNs, the capability of transformer to capture global interactions between elements of a scene without built-in inductive prior along with the formulation of semantic segmentation as a sequence-to-sequence problem facilitates the leverage of contextual information at each stage of the model\cite{strudel2021segmenter}.

\begin{figure}[t]
\centering
\includegraphics[width=\linewidth]{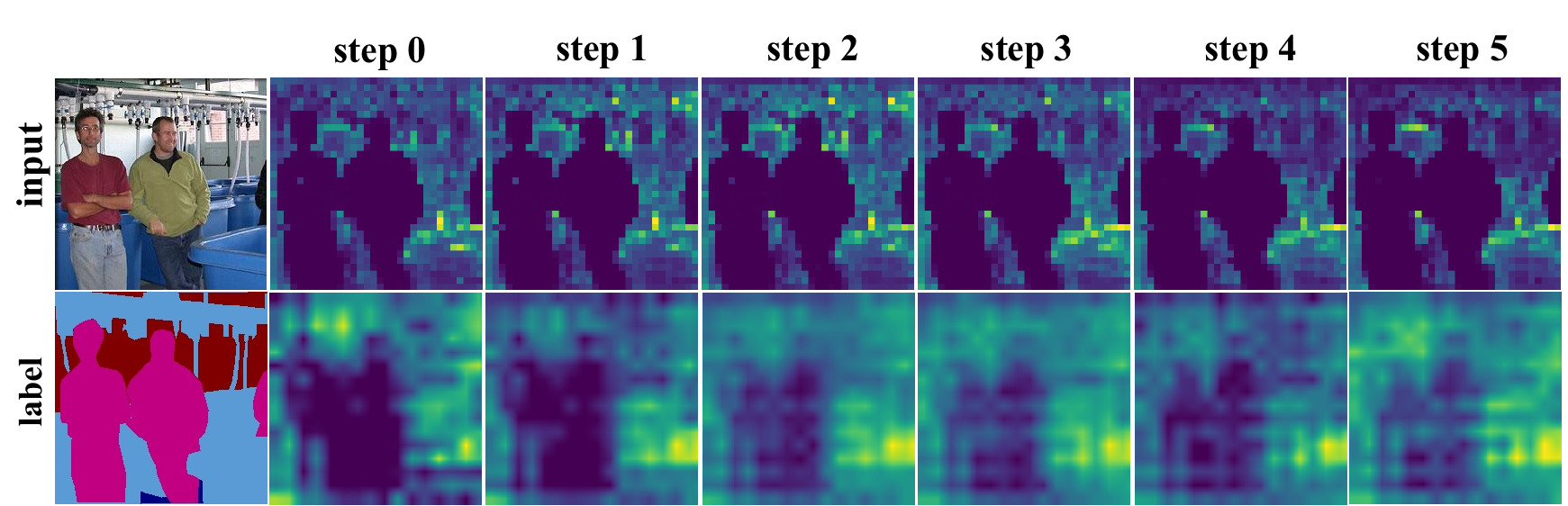}
\vskip -0.3cm
\caption{Comparison between feature maps extracted from \textsl{MiB} method applying Transformer based architecture and CNNs based architecture at different steps of ADE20K \textbf{100-10} task. The first row are the input image and the feature maps extracted from the last layer of encoders in Transformer based architecture at each step. The second row are the corresponding segmentation map and the feature maps extracted from the last layer of encoders in CNNs based architecture at each step.} 
\label{fig:fmap}
\vskip -0.5cm
\end{figure}

As Figure \ref{fig:fmap} shows, we conduct a comparison between the feature maps extracted from \textsl{MiB} method applying Transformer and CNNs based architecture at different steps of a commonly used ISS task ADE20K 100-10(more details are discussed in Experiment Section). Obviously, the feature maps extracted from the Transformer based architecture are regular and continuous, and the information extracted in the previous steps is well preserved, whereas the feature maps extracted from the CNNs based architecture are disordered and the information extracted in the previous steps is lost due to the addition of the new classes.

\subsection{Integration of MiB and Transformer}
We then revisit the distillation based method \textsl{MiB} as it is a classical incremental learning approach tailored to semantic segmentation. As Figure \ref{fig:framework} shows, we follow the unbiased cross entropy loss $\mathcal{L}^t_\text{unce}$(Eq.\ref{eq:unce}) and the unbiased knowledge distillation loss $\mathcal{L}^t_\text{unkd}$(Eq.\ref{eq:unkd}) introduced by \textsl{MiB} to improve the capability of Transformer based architecture to simulate the semantic shift of the background and effectively learn new classes without deteriorating its ability to recognize old ones.

\begin{small}
\begin{equation}
\mathcal{L}^t_\text{unce}(x, y)=-\frac{1}{|\mathcal{I}|} \sum_{i \in \mathcal{I}} \log \tilde{q}_{x}^{t}\left(i, y_{i}\right) 
\label{eq:unce}
\end{equation}
\end{small}

where:
\begin{small}
\begin{equation}
    \tilde{q}_{x}^{t}(i, c)= \begin{cases}\mathcal{M}_{t}(x)[i, c] & \text { if } c \neq \mathrm{b} \\ \sum_{k \in \mathcal{Y}_{t-1}} \mathcal{M}_{t}(x)[i, k] & \text { if } c=\mathrm{b}\end{cases} 
\end{equation}
\end{small}

\begin{small}
\begin{equation}
\mathcal{L}^t_{unkd}(x, y)=-\frac{1}{|\mathcal{I}|} \sum_{i \in \mathcal{I}} \sum_{c \in \mathcal{Y}_{t-1}} \mathcal{M}_{t-1}(x)[i, c] \log \hat{q}_{x}^{t}(i, c) 
\label{eq:unkd}
\end{equation}
\end{small}

where:
\begin{small}
\begin{equation}
\hat{q}_{x}^{t}(i, c)= \begin{cases}\mathcal{M}_{t}(x)[i, c] & \text { if } c \neq \mathrm{b} \\ \sum_{k \in \mathcal{C}_{t}} \mathcal{M}_{t}(x)[i, k] & \text { if } c=\mathrm{b}\end{cases} 
\end{equation}
\end{small}

Considering the Transformer based architecture, a transformer encoder composed of $L$ layers is applied to map the input sequence of embedded patches with position encoding to a sequence of contextualized encodings containing rich semantic information to decode\cite{strudel2021segmenter}. For decoder, we use the linear decoder rather than the mask transformer decoder\cite{strudel2021segmenter} inspired by \textsl{DETR}\cite{carion2020end} since the parameters initialization strategy in \textsl{MiB} method is easier to migrate to linear decoder. We denote the linear decoder parameters as $\left\{\omega_{c}^{t}, \beta_{c}^{t}\right\} \in \theta^{t}$ for the class $c$ at learning step $t$, where $\omega$ and $\beta$ denote its weights and bias. Then the parameters initialization strategy in \textsl{MiB} method is defined as follows.

\begin{small}
\begin{equation}
\omega_{c}^{t}= \begin{cases}\omega_{\mathrm{b}}^{t-1} & \text { if } c \in \mathcal{C}^{t} \\ \omega_{c}^{t-1} & \text { otherwise }\end{cases} 
\end{equation}
\end{small}

\begin{small}
\begin{equation}
\beta_{c}^{t}= \begin{cases}\beta_{\mathrm{b}}^{t-1}-\log \left(\left|\mathcal{C}^{t}\right|\right) & \text { if } c \in \mathcal{C}^{t} \\ \beta_{c}^{t-1} & \text { otherwise }\end{cases} 
\end{equation}
\end{small}

where $\left\{\omega_{c}^{t}, \beta_{c}^{t}\right\}$ are the weights and bias of the background decoder at $t$-$1$ learning step.

\subsection{Patch-wise Losses}

However, when it comes to more challenging multi-step addition tasks, the improvements of Transformer on ISS are relatively limited. To overcome these limitations, we believe that Transformer on ISS can be further enhanced through distilling knowledge at feature space, such as feature imitation. In this section,  we first apply naive $\mathcal{L}^t_{1}$ and $\mathcal{L}^t_{2}$ distillation losses which are commonly used in knowledge distillation on CNN based model between patch representations extracted by Transformer based encoder. Specifically, consider the training step $t (t \geq 1)$, then given an image $\boldsymbol{x}$, $\boldsymbol{h}_t^l=\left\{h_t^l\right\}_{i}$ and $\boldsymbol{h}_{t-1}^l=\left\{h_{t-1}^{l}\right\}_{i}$ denote its patches at the last layer of the model $\mathcal{M}_t$ and the model $\mathcal{M}_{t-1}$, respectively, where $l = 1, 2, \dots, L$ for Transformer based encoder with $L$ layers. The $\mathcal{L}^t_{1}$ and $\mathcal{L}^t_{2}$ distillation losses are defined as follows. 

\begin{small}
\begin{equation}
\mathcal{L}^t_{1} =\frac{1}{L} \sum_{l=1}^{L} \sum_{i} \left|\left\{h_{t}^{l}\right\}_{i}-\left\{h_{t-1}^{l}\right\}_{i}\right|
\end{equation}
\end{small}

\begin{small}
\begin{equation}
\mathcal{L}^t_{2} =\frac{1}{L} \sum_{l=1}^{L} \sum_{i} \left(\left\{h_{t}^{l}\right\}_{i}-\left\{h_{t-1}^{l}\right\}_{i}\right)^2
\end{equation}
\end{small}

However, as shown in Table \ref{table:ablation}, unlike the significant improvement on CNN based model, the direct introduce of $\mathcal{L}^t_{1}$ distillation loss or $\mathcal{L}^t_{2}$ distillation loss degrade the performance of Transformer on ISS. Consequently, to derive a knowledge distillation based method which is more applicable to Transformer based architecture on ISS, we turn to analyse the diversity across patch representations\cite{gong2021vision} extracted from transformer encoder by computing the patch-wise absolute negative and positive cosine similarity to derive how the patch diversification affects the performance on ISS. In detail, when it comes to two sequences of patch representations $\boldsymbol{h}_{1}=\left[\left\{h_{1}\right\}_{1}, \left\{h_{1}\right\}_{2}, \cdots, \left\{h_{1}\right\}_{n}\right]$ and $\boldsymbol{h}_{2}=\left[\left\{h_{2}\right\}_{1}, \left\{h_{2}\right\}_{2}, \cdots, \left\{h_{2}\right\}_{n}\right]$, the patch-wise absolute negative cosine similarity between $\boldsymbol{h}_{1}$ and $\boldsymbol{h}_{2}$ is defined as $\mathcal{S}_\text{negative}(\boldsymbol{h}_{1}, \boldsymbol{h}_{2})$ and the patch-wise absolute positive cosine similarity between $\boldsymbol{h}_{1}$ and $\boldsymbol{h}_{2}$ is defined as $\mathcal{S}_\text{positive}(\boldsymbol{h}_{1}, \boldsymbol{h}_{2})$ as follows.

\begin{small}
\begin{equation}
    \mathcal{P}(h_{i}, h_{j})=\frac{\left|h_{i}^{\top} h_{j}\right|}{\left\|h_{i}\right\|_{2}\left\|h_{j}\right\|_{2}} 
\end{equation}
\end{small}

\begin{small}
\begin{equation}
\mathcal{S}_\text{negative}(\boldsymbol{h}_{1}, \boldsymbol{h}_{2})=\frac{1}{n(n-1)} \sum_{i \neq j} \mathcal{P}(\left\{h_{1}\right\}_{i}, \left\{h_{2}\right\}_{j}) 
\end{equation}
\end{small}

\begin{small}
\begin{equation}
\mathcal{S}_\text{positive}(\boldsymbol{h}_{1}, \boldsymbol{h}_{2})=\frac{1}{n} \sum_{i=1}^{n} \mathcal{P}(\left\{h_{1}\right\}_{i}, \left\{h_{2}\right\}_{i}) 
\end{equation}
\end{small}

First, to access how the relationship between the patch representations extracted from the last layer of both $\mathcal{M}_{t}$ and $\mathcal{M}_{t-1}$ affects the capacity of current model $\mathcal{M}_{t}$ to suppress catastrophic forgetting, we compute the patch-wise negative and positive absolute cosine similarity between $\boldsymbol{h}_t^L$ and $\boldsymbol{h}_{t-1}^L$ at each step of the ADE20K \textbf{100-10} task. Second, the generalization ability of current model $\mathcal{M}_{t}$ on new adding tasks is also crucial, so we compute the patch-wise negative and positive absolute cosine similarity between $\boldsymbol{h}_t^{1}$ and $\boldsymbol{h}_{t}^L$ at each step of the ADE20K \textbf{100-10} task to access how the diversity between the patch representations extracted from the first and the last layer of $\mathcal{M}_{t}$ affects the generalization ability of $\mathcal{M}_{t}$. We conduct the experiments among three methods for incremental learning on Vision Transformer(ViT) based architecture: (\romannumeral1) simple finetuning on the new task $\mathcal{T}^{t}$, denoted as \textsl{ViT-FT}. (\romannumeral2) Traditional distillation based method Learning without forgetting(LwF)\cite{li2017learning}, denoted as \textsl{ViT-LwF}. (\romannumeral3) \textsl{MiB} for ISS\cite{cermelli2020modeling}, denoted as \textsl{ViT-MiB}. 

\begin{figure}[h]
\vskip -0.3cm
    \centering
    \captionsetup{font={small}}
	\subcaptionbox{$\mathcal{S}_\text{positive}(\boldsymbol{h}_t^L, \boldsymbol{h}_{t-1}^L)$}{\includegraphics[width=4.1cm]{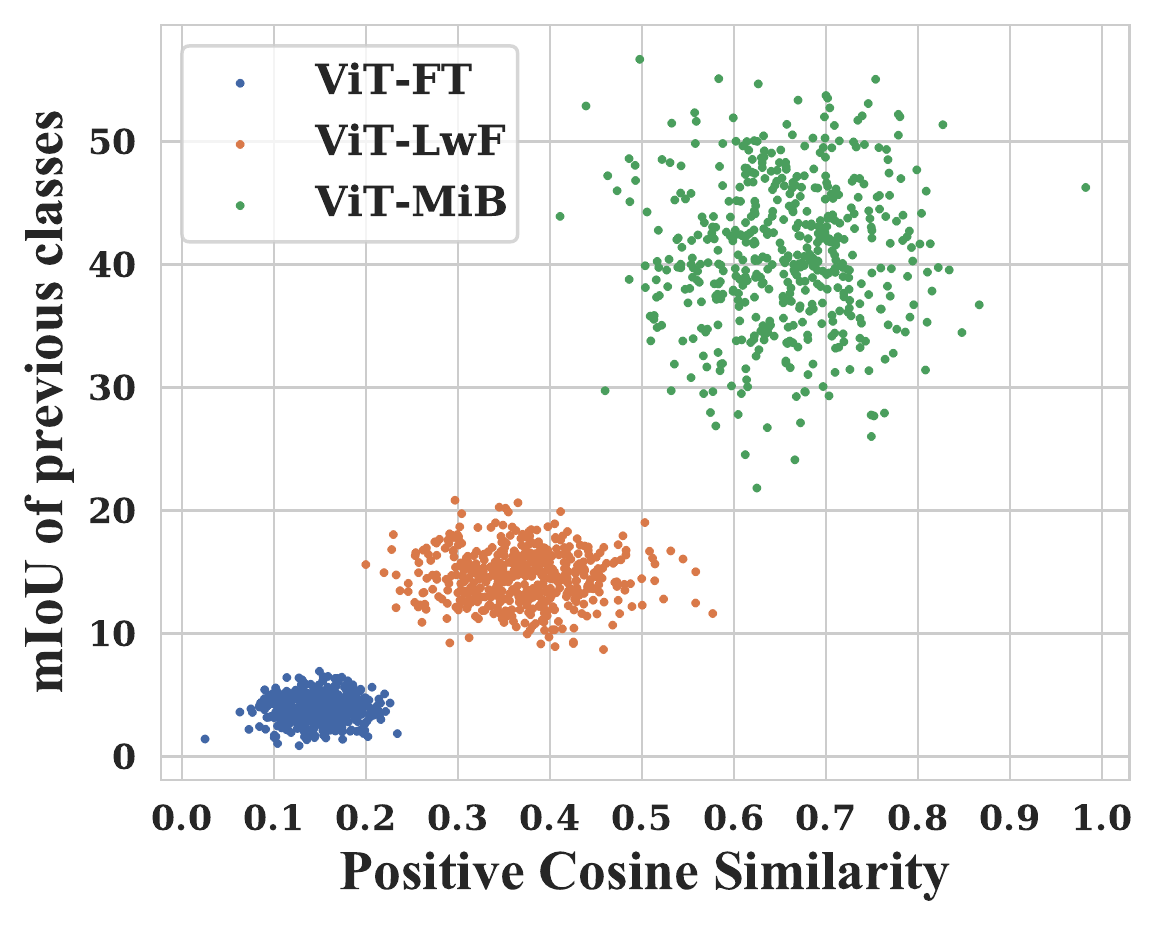}}
	\subcaptionbox{$\mathcal{S}_\text{negative}(\boldsymbol{h}_t^L, \boldsymbol{h}_{t-1}^L)$}{\includegraphics[width=4.1cm]{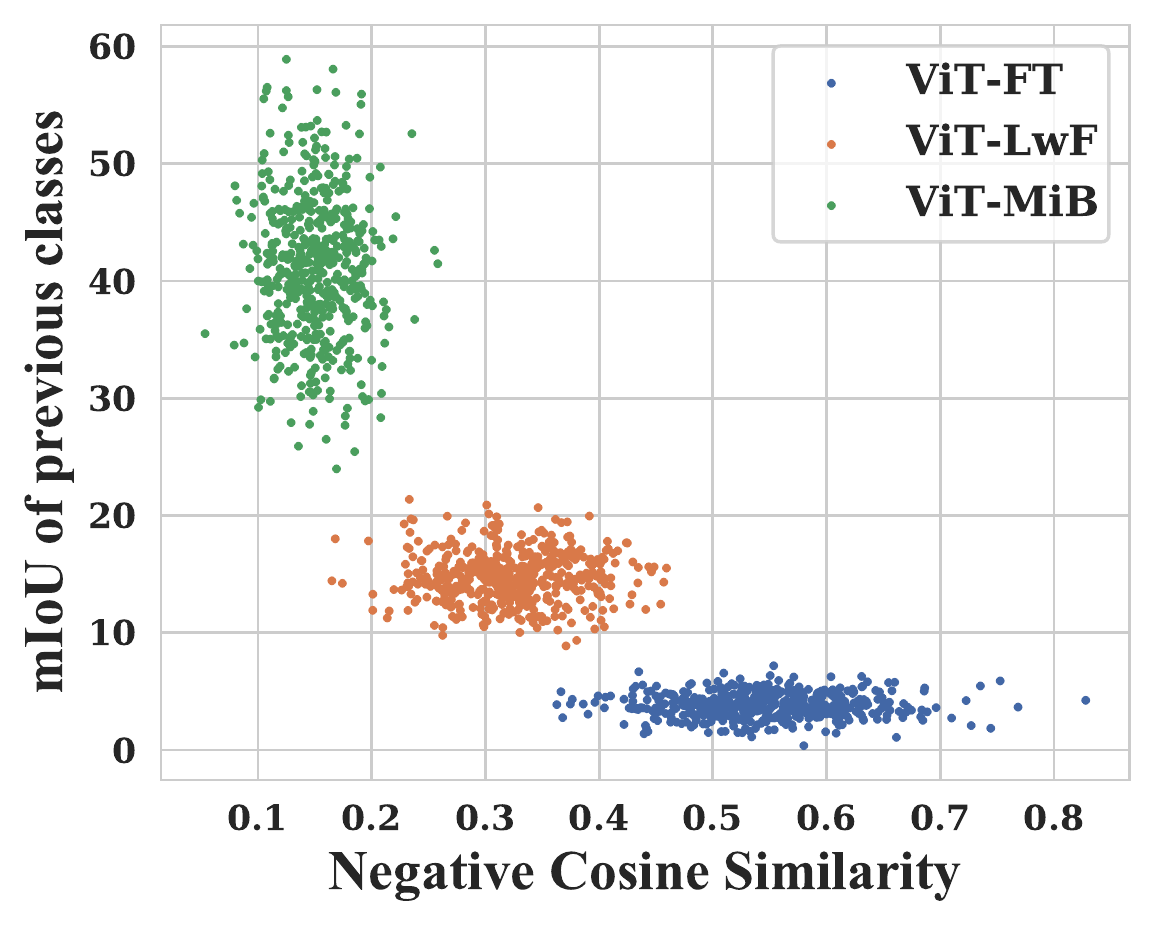}}
	\subcaptionbox{$\mathcal{S}_\text{positive}(\boldsymbol{h}_t^L, \boldsymbol{h}_{t}^1)$}{\includegraphics[width=4.1cm]{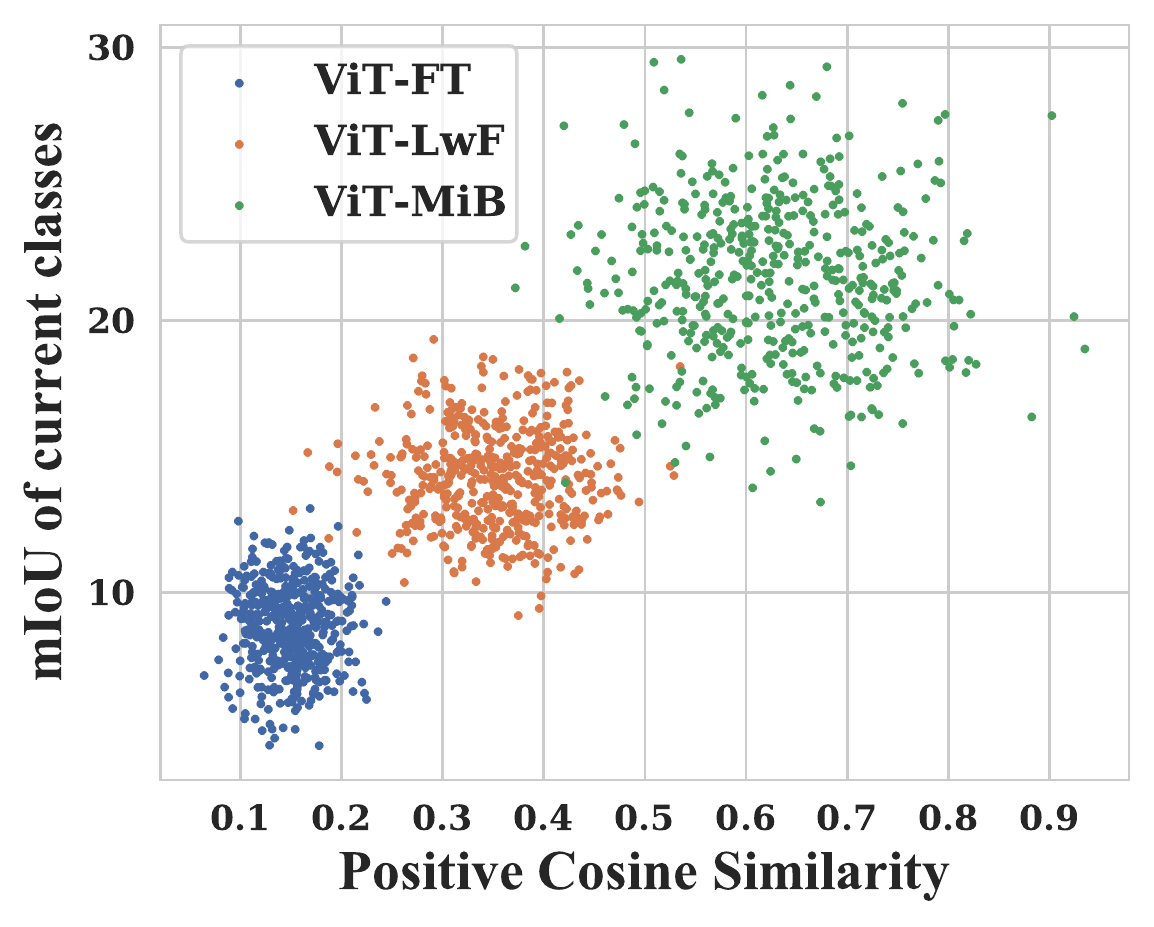}}
	\subcaptionbox{$\mathcal{S}_\text{negative}(\boldsymbol{h}_t^L, \boldsymbol{h}_{t}^1)$}{\includegraphics[width=4.1cm]{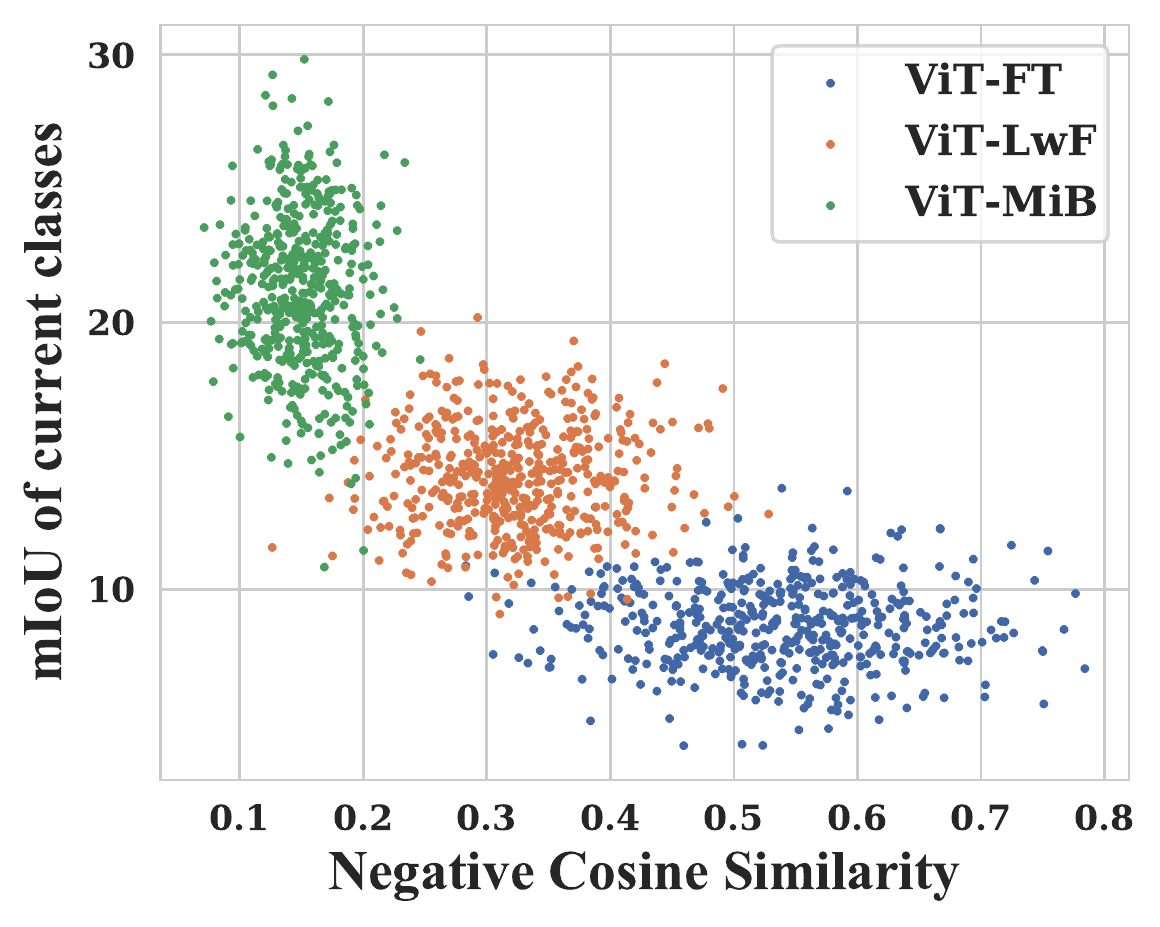}}
	\vskip -0.3cm
	\caption{The first row are the comparisons of patch-wise cosine similarities between $\boldsymbol{h}_t^L$ and $\boldsymbol{h}_{t-1}^L$ along with the corresponding mIoU of previous classes under different methods. The second row are the comparisons of patch-wise cosine similarities between $\boldsymbol{h}_t^{1}$ and $\boldsymbol{h}_t^L$ along with the corresponding mIoU of current classes under different methods. All similarities are computed with 500 images sampled from the ADE20K validation set without data augmentation.}
	\label{fig:losses}
	\vskip -0.5cm
\end{figure}

As depicted in Figure \ref{fig:losses}, methods with relatively high $\mathcal{S}_\text{negative}(\boldsymbol{h}_t^L, \boldsymbol{h}_{t-1}^L)$ and relatively low $\mathcal{S}_\text{positive}(\boldsymbol{h}_t^L, \boldsymbol{h}_{t-1}^L)$ struggle in accumulating the knowledge learned in previous tasks. Meanwhile, methods with lower $\mathcal{S}_\text{negative}(\boldsymbol{h}_t^L, \boldsymbol{h}_{t}^1)$ and higher $\mathcal{S}_\text{positive}(\boldsymbol{h}_t^L, \boldsymbol{h}_{t}^1)$ have better performance on new tasks. In essence, the reasons why \textsl{ViT-MiB} surpasses the other two methods in both old and new tasks are as follows: (\romannumeral 1) the introduction of unbiased knowledge distillation loss $\mathcal{L}^t_{unkd}$ (Eq. \ref{eq:unkd}) fundamentally brings each $\left\{h_{t}^{L}\right\}_{i}$ to be similar to $\left\{h_{t-1}^{L}\right\}_{i}$ and to be different to any other patches $\left\{h_{t-1}^{L}\right\}_{j \neq i}$. (\romannumeral 2) The introduction of unbiased cross entropy loss $\mathcal{L}^t_{unce}$ (Eq. \ref{eq:unce}) essentially ensures the similarity between $\left\{h_{t}^{L}\right\}_{i}$ and $\left\{h_{t}^{1}\right\}_{i}$ while enhancing the patch diversification between $\left\{h_{t}^{L}\right\}_{i}$ and $\left\{h_{t}^{1}\right\}_{j \neq i}$. Based on the above analysis, we propose two patch-wise losses to improve the performance of Transformer on ISS.

\subsubsection{Patch-wise contrastive distillation loss.} Firstly, to enhance the capability of the Transformer based architecture to accumulate knowledge, we propose a patch-wise contrastive distillation loss $\mathcal{L}^t_{\text {cd}}$ between the learned representations from the last layer of $\mathcal{M}_{t-1}$ and $\mathcal{M}_{t}$. We constrain each $\left\{h_{t}^{L}\right\}_{i}$ to be similar to $\left\{h_{t-1}^{L}\right\}_{i}$ and to be different to any other patches $\left\{h_{t-1}^{L}\right\}_{j \neq i}$ as follows. 

\begin{small}
\begin{equation}
    \left\{\mathcal{P}^{kl}_{mn}\right\}_{ij} = \mathcal{P}(\left\{h_{m}^{k}\right\}_{i}, \left\{h_{n}^{l}\right\}_{j})
\end{equation}
\end{small}

\begin{small}
\begin{equation}
\mathcal{L}^t_{\text {cd}}=-\frac{1}{n} \sum_{i=1}^{n} \log \frac{\exp \left(\left\{\mathcal{P}^{LL}_{t(t-1)}\right\}_{ii}\right)}{\exp \left(\left\{\mathcal{P}^{LL}_{t(t-1)}\right\}_{ii}\right)+\sum_{j \neq i}\exp \left(\left\{\mathcal{P}^{LL}_{t(t-1)}\right\}_{ij}\right)}  
\end{equation}
\end{small}

\subsubsection{Patch-wise contrastive loss.} Secondly, to improve the generalization ablity of the Transformer based architecture on new tasks, we propose a patch-wise contrastive loss $\mathcal{L}^t_{\text {ct}}$ between the learned representations from the first layer and the last layer of the model at step $t$. We constrain each $\left\{h_t^L\right\}_{i}$ to be similar to $\left\{h_t^{1}\right\}_{i}$ and to be different to any other patches $\left\{h_t^{1}\right\}_{j \neq i}$ as follows. 

\begin{small}
\begin{equation}
\mathcal{L}^t_{\text {ct}}=-\frac{1}{n} \sum_{i=1}^{n} \log \frac{\exp \left(\left\{\mathcal{P}^{L1}_{tt}\right\}_{ii}\right)}{\exp \left(\left\{\mathcal{P}^{L1}_{tt}\right\}_{ii}\right)+\sum_{j \neq i}\exp \left(\left\{\mathcal{P}^{L1}_{tt}\right\}_{ij}\right)} 
\end{equation}
\end{small}

To conclude, the total loss at step $t$ is defined as $\mathcal{L}^{t}_{\text {total}}$, where $w^{t}_{\left(\cdot\right)}$ is the corresponding weight.

\begin{small}
\begin{equation}
\mathcal{L}^{t}_{\text {total}}= \begin{cases} w^{t}_{\text{unce}}\mathcal{L}^t_{\text{unce}}+w^{t}_{\text{ct}}\mathcal{L}^t_{\text{ct}} & \mathrm{t}=0  \\ w^{t}_{\text{unce}}\mathcal{L}^t_{\text{unce}}+w^{t}_{\text{unkd}}\mathcal{L}^t_{\text{unkd}}+w^{t}_{\text{cd}}\mathcal{L}^t_{\text{cd}}+w^{t}_{\text{ct}}\mathcal{L}^t_{\text{ct}} & \mathrm{t} \geq 1\end{cases}
\end{equation}
\end{small}

\section{Experiments}

\begin{table*}[t]
\centering
\captionsetup{font={small}}
    \caption{mIoU on the Pascal-VOC 2012 dataset for different incremental class learning scenarios. Best in \textbf{bold}, runner-up {\ul underlined}. $\dagger$: results from \cite{cermelli2020modeling}, $\diamond$: results from \cite{michieli2021continual}, $\star$: results from \cite{douillard2021tackling}. $\ddag$: all BN layers are frozen after the first training step.}
    \label{table:voc}
    \vskip -0.3cm
\resizebox{\textwidth}{!}{
\begin{tabular}{lllcccccc||cccccc||cccccc}
\hline
                                  &            &            & \multicolumn{6}{||c||}{\textbf{19-1}}                                                                                                                                    & \multicolumn{6}{c||}{\textbf{15-5}}                                                                                                                                    & \multicolumn{6}{c}{\textbf{15-1}}                                                                                                                           \\\hline
                                  &            &            & \multicolumn{3}{||c||}{\textbf{Disjoint}}                                                           & \multicolumn{3}{c||}{\textbf{Overlapped}}                                    & \multicolumn{3}{c||}{\textbf{Disjoint}}                                                           & \multicolumn{3}{c||}{\textbf{Overlapped}}                                    & \multicolumn{3}{c||}{\textbf{Disjoint}}                                                 & \multicolumn{3}{c}{\textbf{Overlapped}}                                     \\\hline
\multicolumn{1}{l||}{\textbf{Method}} & \multicolumn{1}{l||}{\textbf{Architecture}} & \textbf{Backbone}   & \multicolumn{1}{||c}{\textsl{1-19}}          & \multicolumn{1}{c|}{\textsl{20}}            & \multicolumn{1}{c||}{\textsl{all}}           & \textsl{1-19}          & \multicolumn{1}{c|}{\textsl{20}}            & \textsl{all}           & \textsl{1-15}       & \multicolumn{1}{c|}{\textsl{16-20}}         & \multicolumn{1}{c||}{\textsl{all}}           & \textsl{1-15}          & \multicolumn{1}{c|}{\textsl{16-20}}         & \textsl{all}           & \textsl{1-15}          & \multicolumn{1}{c|}{\textsl{16-20}}         & \multicolumn{1}{c||}{\textsl{all}} & \textsl{1-15}          & \multicolumn{1}{c|}{\textsl{16-20}}         & \textsl{all}           \\ \hline
\multicolumn{1}{l||}{FT$\dagger$} &   \multicolumn{1}{l||}{Deeplab-v3}            & ResNet-101 & \multicolumn{1}{||c}{5.8}           & \multicolumn{1}{c|}{12.3}          & \multicolumn{1}{c||}{6.2}           & 6.8           & \multicolumn{1}{c|}{12.9}          & 7.1           & 1.1           & \multicolumn{1}{c|}{33.6}          & \multicolumn{1}{c||}{9.2}           & 2.1           & \multicolumn{1}{c|}{33.1}          & 9.8           & 0.2           & \multicolumn{1}{c|}{1.8}           & \multicolumn{1}{c||}{0.6}                      & 0.2           & \multicolumn{1}{c|}{1.8}           & 0.6           \\
\multicolumn{1}{l||}{FT$\diamond$} &   \multicolumn{1}{l||}{Deeplab-v3+}            & ResNet-101 & \multicolumn{1}{||c}{35.2}           & \multicolumn{1}{c|}{13.2}          & \multicolumn{1}{c||}{34.2}           & 34.7           & \multicolumn{1}{c|}{14.9}          & 33.8           & 8.4           & \multicolumn{1}{c|}{33.5}          & \multicolumn{1}{c||}{14.4}           & 12.5           & \multicolumn{1}{c|}{36.9}          & 18.3          & 5.8           & \multicolumn{1}{c|}{4.9}           & \multicolumn{1}{c||}{5.6}                      & 4.9           & \multicolumn{1}{c|}{3.2}           & 4.5           \\
\multicolumn{1}{l||}{FT} &   \multicolumn{1}{l||}{Transformer}           & ViT-small   & \multicolumn{1}{||c}{5.9}           & \multicolumn{1}{c|}{50.9}          & \multicolumn{1}{c||}{8.1}          & 6.0             & \multicolumn{1}{c|}{53.1}          & 8.4          & 17.9          & \multicolumn{1}{c|}{45.8}          & \multicolumn{1}{c||}{24.9}          & 19.3            & \multicolumn{1}{c|}{49.1}          & 26.8          & 4.6           & \multicolumn{1}{c|}{35.8}          & \multicolumn{1}{c||}{12.4}                     & 4.7          & \multicolumn{1}{c|}{38.3}          & 13.1          \\ \hline
\multicolumn{1}{l||}{SDR$\diamond$} &   \multicolumn{1}{l||}{Deeplab-v3+}          & ResNet-101 & \multicolumn{1}{||c}{69.9}          & \multicolumn{1}{c|}{37.3}          & \multicolumn{1}{c||}{68.4}          & 69.1          & \multicolumn{1}{c|}{32.6}          & 67.4          & 73.5          & \multicolumn{1}{c|}{47.3}          & \multicolumn{1}{c||}{67.2}          & 75.4          & \multicolumn{1}{c|}{52.6}          & 69.9          & 59.2          & \multicolumn{1}{c|}{12.9}          & \multicolumn{1}{c||}{48.1}                     & 44.7          & \multicolumn{1}{c|}{21.8}          & 39.2          \\
\multicolumn{1}{l||}{SDR+MiB$\diamond$}  &   \multicolumn{1}{l||}{Deeplab-v3+}         & ResNet-101 & \multicolumn{1}{||c}{70.8}          & \multicolumn{1}{c|}{31.4}          & \multicolumn{1}{c||}{68.9}          & 71.3          & \multicolumn{1}{c|}{23.4}          & 69.0          & 74.6          & \multicolumn{1}{c|}{44.1}          & \multicolumn{1}{c||}{67.3}          & 76.3         & \multicolumn{1}{c|}{50.2}          & 70.1          &  59.4         & \multicolumn{1}{c|}{14.3}          & \multicolumn{1}{c||}{48.7}                     & 47.3          & \multicolumn{1}{c|}{14.7}          & 39.5          \\
\multicolumn{1}{l||}{PLOP$\star$} &   \multicolumn{1}{l||}{Deeplab-v3}          & ResNet-101 & \multicolumn{1}{||c}{-}             & \multicolumn{1}{c|}{-}             & \multicolumn{1}{c||}{-}             & {\ul 75.4}    & \multicolumn{1}{c|}{{\ul 37.4}}    & {\ul 73.5}    & -             & \multicolumn{1}{c|}{-}             & \multicolumn{1}{c||}{-}             & 75.7    & \multicolumn{1}{c|}{51.7}          & 70.1    & -             & \multicolumn{1}{c|}{-}             & \multicolumn{1}{c||}{-}                        & {\ul 65.1}    & \multicolumn{1}{c|}{21.1}    & 54.6   \\
\multicolumn{1}{l||}{MiB$\dagger$}  &   \multicolumn{1}{l||}{Deeplab-v3}          & ResNet-101 & \multicolumn{1}{||c}{69.6}         & \multicolumn{1}{c|}{25.6}          & \multicolumn{1}{c||}{67.4}          & 70.2          & \multicolumn{1}{c|}{22.1}          & 67.8          & 71.8          & \multicolumn{1}{c|}{43.3}          & \multicolumn{1}{c||}{64.7}          & 75.5          & \multicolumn{1}{c|}{49.4}          & 69.0          & 46.2          & \multicolumn{1}{c|}{12.9}          & \multicolumn{1}{c||}{37.9}                     & 35.1          & \multicolumn{1}{c|}{13.5}          & 29.7          \\
\multicolumn{1}{l||}{MiB$\ddag$}  &   \multicolumn{1}{l||}{Deeplab-v3}          & ResNet-101 & \multicolumn{1}{||c}{71.3}         & \multicolumn{1}{c|}{23.6}          & \multicolumn{1}{c||}{68.9}          & 72.4          & \multicolumn{1}{c|}{20.7}          & 69.8          & 73.6          & \multicolumn{1}{c|}{41.2}          & \multicolumn{1}{c||}{65.5}          & 78.2          & \multicolumn{1}{c|}{45.7}          & 70.1          & 49.4          & \multicolumn{1}{c|}{9.8}          & \multicolumn{1}{c||}{39.5}                     & 38.7          & \multicolumn{1}{c|}{10.3}          & 31.6          \\
\multicolumn{1}{l||}{MiB$\diamond$}  &   \multicolumn{1}{l||}{Deeplab-v3+}          & ResNet-101 & \multicolumn{1}{||c}{67.0}         & \multicolumn{1}{c|}{26.0}          & \multicolumn{1}{c||}{65.1}          & 69.6          & \multicolumn{1}{c|}{23.8}          & 67.4          & 47.5          & \multicolumn{1}{c|}{34.1}          & \multicolumn{1}{c||}{44.3}          & 73.1          & \multicolumn{1}{c|}{44.5}          & 66.3          & 39.0          & \multicolumn{1}{c|}{15.0}          & \multicolumn{1}{c||}{33.3}                     & 44.5          & \multicolumn{1}{c|}{11.7}          & 36.7          \\
\multicolumn{1}{l||}{MiB}  &   \multicolumn{1}{l||}{Transformer}          & ViT-small & \multicolumn{1}{||c}{{\ul 74.3}}         & \multicolumn{1}{c|}{\ul 37.6}          & \multicolumn{1}{c||}{\ul 73.8}          & 74.9          & \multicolumn{1}{c|}{37.2}          & 73.1          & {\ul 76.9}         & \multicolumn{1}{c|}{\ul 50.7}          & \multicolumn{1}{c||}{\ul 70.4}          & {\ul 77.4}          & \multicolumn{1}{c|}{\ul 53.4}          & {\ul 70.9}          & {\ul 68.4}          & \multicolumn{1}{c|}{\ul 35.8}          & \multicolumn{1}{c||}{\ul 60.7}                     & 64.9          & \multicolumn{1}{c|}{\ul 30.7}          & {\ul 60.4}          \\
\multicolumn{1}{l||}{TISS(ours)} &   \multicolumn{1}{l||}{Transformer}   & ViT-small   & \multicolumn{1}{||c}{\textbf{81.8}} & \multicolumn{1}{c|}{\textbf{50.8}} & \multicolumn{1}{c||}{\textbf{80.3}} & \textbf{81.5} & \multicolumn{1}{c|}{\textbf{49.8}} & \textbf{79.9} & \textbf{80.9} & \multicolumn{1}{c|}{\textbf{61.7}} & \multicolumn{1}{c||}{\textbf{77.3}} & \textbf{81.9} & \multicolumn{1}{c|}{\textbf{67.8}} & \textbf{78.5} & \textbf{77.3} & \multicolumn{1}{c|}{\textbf{45.6}} & \multicolumn{1}{c||}{\textbf{69.4} }           & \textbf{78.9} & \multicolumn{1}{c|}{\textbf{51.7}} & \textbf{72.1} \\ \hline
\multicolumn{1}{l||}{offline$\dagger$} &   \multicolumn{1}{l||}{Deeplab-v3}       & ResNet-101 & \multicolumn{1}{||c}{77.4}          & \multicolumn{1}{c|}{78.0}          & \multicolumn{1}{c||}{77.4}          & 77.4          & \multicolumn{1}{c|}{78.0}          & 77.4          & 79.1          & \multicolumn{1}{c|}{72.6}          & \multicolumn{1}{c||}{77.4}          & 79.1          & \multicolumn{1}{c|}{72.6}          & 77.4          & 79.1          & \multicolumn{1}{c|}{72.6}          & \multicolumn{1}{c||}{77.4}                    & 79.1          & \multicolumn{1}{c|}{72.6}          & 77.4          \\
\multicolumn{1}{l||}{offline$\diamond$} &   \multicolumn{1}{l||}{Deeplab-v3+}       & ResNet-101 & \multicolumn{1}{||c}{78.9}          & \multicolumn{1}{c|}{78.1}          & \multicolumn{1}{c||}{78.5}          & 78.9          & \multicolumn{1}{c|}{78.1}          & 78.5         & 79.9          & \multicolumn{1}{c|}{73.1}          & \multicolumn{1}{c||}{78.5}          & 79.9          & \multicolumn{1}{c|}{73.1}          & 78.5          & 79.9          & \multicolumn{1}{c|}{73.1}          & \multicolumn{1}{c||}{78.5}                    & 79.9          & \multicolumn{1}{c|}{73.1}          & 78.5          \\
\multicolumn{1}{l||}{offline} &   \multicolumn{1}{l||}{Transformer}      & ViT-small   & \multicolumn{1}{||c}{82.8}          & \multicolumn{1}{c|}{79.2}          & \multicolumn{1}{c||}{81.3}          & 82.8          & \multicolumn{1}{c|}{79.2}          & 81.3          & 84.1          & \multicolumn{1}{c|}{74.3}          & \multicolumn{1}{c||}{81.3}          & 84.1          & \multicolumn{1}{c|}{74.3}          & 81.3          & 84.1          & \multicolumn{1}{c|}{74.3}          & \multicolumn{1}{c||}{81.3}                     & 84.1          & \multicolumn{1}{c|}{74.3}          & 81.3         \\
\hline
\end{tabular}}
\vskip -0.5cm
\end{table*}

We evaluate the performance of \textsl{TISS} against some state-of-the art methods. In all tables we mark the architecture and backbone applied by all methods. Specifically, we compare \textsl{TISS} with the following four recent ISS methods: \textsl{MiB}\cite{cermelli2020modeling}, \textsl{SDR}\cite{michieli2021continual}, \textsl{SDR+MiB}\cite{michieli2021continual}, \textsl{PLOP}\cite{douillard2021tackling}. As for metrics, we report Intersection-over-Union(mIoU) over all the classes of one learning step and all the steps.

\subsection{Implementation Details}
\subsubsection{Transformer based architecture.} 
For the encoder shown in Figure \ref{fig:framework}, we build \textsl{TISS} upon the ViT with ``Tiny''(with 6 million parameters), ``Small''(with 22 million parameters),``Base''(with 86 million parameters) and ``Large''(with 307 million parameters) models. Compared to Deeplab-v3 architecture with a ResNet-101\cite{he2016deep} backbone(with 58 million parameters), we use the ViT-small backbone as encoder in all experiments due to the fewer parameters and competitive performance. For decoder, we use the linear decoder to map the patch representations to segmentation map.

\subsubsection{Training configuration.}
We train \textsl{TISS} with SGD optimizer and the same learning rate policy, momentum and weight decay. We set the initial learning rate to $10^{-2}$ for the first learning step and decrease it to $10^{-3}$ for the following steps as done in \cite{cermelli2020modeling}. The learning rate is decreased with a polynomial strategy with power $0.9$. The batch size is 12 with 30 epochs of training for Pascal-VOC 2012. For ADE20K, batch size is set to 8 and epoch is set to 60. We apply the same data augmentation of \cite{chen2017rethinking} and crop the images to $512 \times 512$ during both training and validation. The hyper-parameters of each method are set refer to the protocol of incremental learning defined in \cite{de2019continual}, using $20 \%$ of the training set as validation. The final results are reported on the standard validation set of the datasets. We set the $w^t_{unce}$, $w^t_{ct}$ and $w^t_{cd}$ to $1.0$, $0.1$ and $0.1$ for each step at all tasks, respectively. Then we set $w^t_{unkd}$ to $10.0$ for the tasks of which the new classes are added at once and to $30.0$ for the tasks of which the new classes are added sequentially.

\subsection{Pascal-VOC 2012}
Pascal-VOC 2012 is a widely used benchmark consists of 10582 images in the training split and 1449 in the validation split with 20 foreground object classes. Following previous work\cite{cermelli2020modeling,michieli2019incremental}, two experimental protocals are defined as \textsl{disjoint} setup and \textsl{overlapped} setup. The \textsl{disjoint} setup assumes that each learning step contains a unique set of images whose pixels belong to classes seen either in the current or in the previous learning steps. With this setup, only pixels of novel classes are labelled, while the old ones are labeled as background. The \textsl{overlapped} setup assumes that each learning step contains all the images which have at least one pixel of a novel class with only the latter annotated. With this setup, training set contains images with pixels of classes that we will learn in the future, but labeled as background in the ground truth. We conduct three different experiments involving the addition of one class(\textbf{19-1}), five classes all at once(\textbf{15-5}), and five classes sequentially(\textbf{15-1}) with these two setups. In Table \ref{table:voc} we present comprehensive results on the experiments above with both \textsl{disjoint} and \textsl{overlapped} setups. We report the average mIoU of classes in previous step, incremental step and all classes, respectively.

We start by addition of one class(\textbf{19-1}) task. As reported in Table \ref{table:voc}, suffered from catastrophic forgetting, all \textsl{FT} methods perform poorly on all classes. 
In \textsl{MiB} method, Transformer based architecture with ViT-small as backbone(denoted as ${MiB}_{T}$) achieves an average improvement of about $9.8\%$ in mIoU against CNNs based architecture applying the Deeplab-v3 and Deeplab-v3+ architecture with ResNet-101 as backbone(denoted as $MiB_{D}$), which demonstrates that Transformer based architecture is advantageous to ISS. In addition, \textsl{TISS} improves the performance over $MiB_{T}$ by $9.1\%$ in both the \textsl{disjoint} and \textsl{overlapped} scenarios, which reinforces the effectiveness of $\mathcal{L}_\text{cd}$ and $\mathcal{L}_\text{ct}$. Furthermore, \textsl{TISS} surpasses the current state-of-the-art \textsl{PLOP} by nearly $8.7\%$ in the \textsl{overlapped} scenario.

Then we move to single-step addition of five classes(\textbf{15-5}). Results are reported in Table \ref{table:voc}. Overall, the performance of all \textsl{FT} methods are consistent with the \textbf{19-1} setting. Obviously, $MiB_{T}$ leads to a significant boost on the task in the \textsl{disjoint} scenario, which even achieves the competitive performance against \textsl{PLOP} in the \textsl{overlapped} scenario. Meanwhile, \textsl{TISS} achieves an average improvement over the best baseline of $5.5\%$ on old classes, of $25.0\%$ on new classes and $10.3\%$ on all classes in both the \textsl{disjoint} and \textsl{overlapped} scenarios.

In the final scenario we conduct multi-step addition of five classes(more challenging \textbf{15-1} task). As shown in Table \ref{table:voc}, the performance of all existing methods drop significantly compared with previous tasks. It is important to note that all \textsl{FT} methods are difficult to prevent forgetting. Compared with the previous \textbf{15-5} task, even \textsl{PLOP} suffer from a drop of nearly $14.0\%$ on old classes, of $59.2\%$ on new classes and of $22.1\%$ on all classes in the \textsl{overlapped} scenario. Meanwhile, although the performance of $MiB_{T}$ drops significantly compared with offline training, Transformer based architecture still achieves competitive results against \textsl{PLOP} in the \textsl{overlapped} scenario, which reinforces that the Transformer is more applicable to ISS. Furthermore, \textsl{TISS} achieves an average improvement over the best baseline of $25.5\%$ on old classes, of $73.0\%$ on new classes and $30.6\%$ on all classes in both \textsl{disjoint} and \textsl{overlapped} scenarios.

\begin{table*}[t]
\centering
\captionsetup{font={small}}
    \caption{mIoU on the ADE20K dataset for different incremental class learning scenarios. Best in \textbf{bold}, runner-up {\ul underlined}. $\dagger$: results from \cite{cermelli2020modeling}, $\diamond$: results from \cite{michieli2021continual}, $\star$: results from \cite{douillard2021tackling}. $\ddag$: all BN layers are frozen after the first training step.}
    \label{table:ade}
    \vskip -0.3cm
\resizebox{\textwidth}{!}{
\begin{tabular}{lllccc||cccccccc||ccccc}
\hline
                                  &            &            & \multicolumn{3}{||c||}{\textbf{100-50}}                                   & \multicolumn{8}{c||}{\textbf{100-10}}                                                                                                                                                       & \multicolumn{5}{c}{\textbf{50-50}}                                                                                                                 \\\hline
\multicolumn{1}{l||}{\textbf{Method}} & \multicolumn{1}{l||}{\textbf{Architecture}} & \textbf{Backbone}   & \multicolumn{1}{||c}{\textsl{1-100}} & \multicolumn{1}{c|}{\textsl{101-150}} & \textsl{all}           & \textsl{1-100} & \textsl{101-110} & \textsl{111-120} & \textsl{121-130} & \textsl{131-140} & \multicolumn{1}{l|}{\textsl{141-150}} & \multicolumn{1}{l|}{\textsl{101-150}} & \textsl{all}  & \multicolumn{1}{l}{\textsl{1-50}} & \textsl{51-100} & \multicolumn{1}{c|}{\textsl{101-150}} & \multicolumn{1}{l|}{\textsl{51-150}} & \textsl{all}  \\ \hline
\multicolumn{1}{l||}{FT$\dagger$} &   \multicolumn{1}{l||}{Deeplab-v3}         & ResNet-101 & \multicolumn{1}{||c}{0.0}            & \multicolumn{1}{c|}{24.9}             & 8.3           & 0.0            & 0.0              & 0.0              & 0.0              & 0.0              & \multicolumn{1}{c|}{16.6}             & \multicolumn{1}{c|}{4.8}              & 1.1           & 0.0                               & 0.0             & \multicolumn{1}{c|}{22.0}             & \multicolumn{1}{c|}{5.7}             & 0.6           \\
\multicolumn{1}{l||}{FT$\diamond$} &   \multicolumn{1}{l||}{Deeplab-v3+}         & ResNet-101 & \multicolumn{1}{||c}{0.0}            & \multicolumn{1}{c|}{22.5}             & 7.5           & 0.0            & -             & -              & -              & -              & \multicolumn{1}{c|}{-}             & \multicolumn{1}{c|}{2.5}              & 0.8           & 13.9                               & -             & \multicolumn{1}{c|}{-}             & \multicolumn{1}{c|}{12.0}             & 12.6           \\
\multicolumn{1}{l||}{FT}&   \multicolumn{1}{l||}{Transformer}           & ViT-small   & \multicolumn{1}{||c}{1.2}            & \multicolumn{1}{c|}{28.8}             & 10.4          & 0.3            & 0.8              & 0.8              & 1.0              & 1.2              & \multicolumn{1}{c|}{39.7}             & \multicolumn{1}{c|}{8.7}              & 3.1           & 0.5                               & 2.1             & \multicolumn{1}{c|}{35.4}             & \multicolumn{1}{c|}{18.8}            & 12.7           \\ \hline
\multicolumn{1}{l||}{SDR$\diamond$} &   \multicolumn{1}{l||}{Deeplab-v3+}          & ResNet-101 & \multicolumn{1}{||c}{37.4}           & \multicolumn{1}{c|}{24.8}             & 33.2          & 28.9           & -                & -                & -                & -                & \multicolumn{1}{c|}{-}                & \multicolumn{1}{c|}{7.4}              & 21.7          & 40.9                              & -               & \multicolumn{1}{c|}{-}                & \multicolumn{1}{c|}{23.9}             & 29.5          \\
\multicolumn{1}{l||}{SDR+MiB$\diamond$} &   \multicolumn{1}{l||}{Deeplab-v3+}          & ResNet-101 & \multicolumn{1}{||c}{37.5}           & \multicolumn{1}{c|}{25.5}             & 33.5          & 28.9           & -                & -                & -                & -                & \multicolumn{1}{c|}{-}                & \multicolumn{1}{c|}{11.7}              & 23.2          & 42.9                              & -               & \multicolumn{1}{c|}{-}                & \multicolumn{1}{c|}{25.4}             & 31.3          \\
\multicolumn{1}{l||}{PLOP$\star$}&   \multicolumn{1}{l||}{Deeplab-v3}         & ResNet-101 & \multicolumn{1}{||c}{{\ul 41.9}}     & \multicolumn{1}{c|}{14.9}             & 32.9          & {\ul 40.5}     & -                & -                & -                & -                & \multicolumn{1}{c|}{-}                & \multicolumn{1}{c|}{13.6}             & {\ul 31.6}    & {\ul 48.8}                        & -               & \multicolumn{1}{c|}{-}                & \multicolumn{1}{c|}{21.0}            & 30.4   \\
\multicolumn{1}{l||}{MiB$\dagger$}&   \multicolumn{1}{l||}{Deeplab-v3}          & ResNet-101 & \multicolumn{1}{||c}{37.9}           & \multicolumn{1}{c|}{27.9}             & 34.6          & 31.8           & 10.4             & 14.8             & 12.8             & 13.6             & \multicolumn{1}{c|}{{\ul 18.7}}       & \multicolumn{1}{c|}{14.1}                & 25.9          & 35.5                              & 22.2            & \multicolumn{1}{c|}{23.6}             & \multicolumn{1}{c|}{22.9}               & 27.0          \\
\multicolumn{1}{l||}{MiB$\ddag$}&   \multicolumn{1}{l||}{Deeplab-v3}          & ResNet-101 & \multicolumn{1}{||c}{38.7}           & \multicolumn{1}{c|}{26.3}             & 34.6          & 32.9           & 9.2             & 14.1             & 11.6             & 12.3             & \multicolumn{1}{c|}{{17.0}}       & \multicolumn{1}{c|}{12.8}                & 26.2          & 37.3                              & 21.1            & \multicolumn{1}{c|}{21.5}             & \multicolumn{1}{c|}{21.3}               & 26.6          \\
\multicolumn{1}{l||}{MiB$\diamond$}&   \multicolumn{1}{l||}{Deeplab-v3+}          & ResNet-101 & \multicolumn{1}{||c}{37.6}           & \multicolumn{1}{c|}{24.7}             & 33.3          & 21.0           & -             & -            & -             & -             & \multicolumn{1}{c|}{-}       & \multicolumn{1}{c|}{5.3}                & 15.8          & 39.1                              & -            & \multicolumn{1}{c|}{-}             & \multicolumn{1}{c|}{22.6}               & 28.1          \\
\multicolumn{1}{l||}{MiB}&   \multicolumn{1}{l||}{Transformer}          & ViT-small   & \multicolumn{1}{||c}{39.2}           & \multicolumn{1}{c|}{{\ul 28.7}}       & {\ul 35.7}    & 35.4           & {\ul 19.8}       & {\ul 26.4}       & {\ul 29.9}       & {\ul 14.8}       & \multicolumn{1}{c|}{15.3}             & \multicolumn{1}{c|}{{\ul 21.2}}       & 30.7          & 47.8                              & {\ul 33.9}      & \multicolumn{1}{c|}{{\ul 28.1}}       & \multicolumn{1}{c|}{{\ul 31.0}}      & {\ul 36.6}          \\
\multicolumn{1}{l||}{TISS(ours)}&   \multicolumn{1}{l||}{Transformer}   & ViT-small   & \multicolumn{1}{||c}{\textbf{43.0}}  & \multicolumn{1}{c|}{\textbf{29.4}}    & \textbf{38.4} & \textbf{41.2}  & \textbf{21.1}    & \textbf{29.3}    & \textbf{30.6}    & \textbf{15.0}    & \multicolumn{1}{c|}{\textbf{19.8}}    & \multicolumn{1}{c|}{\textbf{23.2}}    & \textbf{35.2} & \textbf{49.7}                     & \textbf{35.1}   & \multicolumn{1}{c|}{\textbf{30.2}}    & \multicolumn{1}{c|}{\textbf{32.7}}   & \textbf{38.3} \\ \hline
\multicolumn{1}{l||}{offline$\dagger$}&   \multicolumn{1}{l||}{Deeplab-v3}      & ResNet-101 & \multicolumn{1}{||c}{44.3}           & \multicolumn{1}{c|}{28.2}             & 38.9          & 44.3           & 26.1             & 42.8             & 26.7             & 28.1             & \multicolumn{1}{c|}{17.3}             & \multicolumn{1}{c|}{28.2}             & 38.9          & 51.1                              & 37.4            & \multicolumn{1}{c|}{28.2}             & \multicolumn{1}{c|}{32.8}            & 38.9          \\
\multicolumn{1}{l||}{offline$\diamond$}&   \multicolumn{1}{l||}{Deeplab-v3+}      & ResNet-101 & \multicolumn{1}{||c}{43.9}           & \multicolumn{1}{c|}{27.2}             & 38.3          & 43.9           & -             & -             & -             & -             & \multicolumn{1}{c|}{-}             & \multicolumn{1}{c|}{27.2}             & 38.3          & 50.9                             & -            & \multicolumn{1}{c|}{-}             & \multicolumn{1}{c|}{32.1}            & 38.3          \\
\multicolumn{1}{l||}{offline}&   \multicolumn{1}{l||}{Transformer}      & ViT-small   & \multicolumn{1}{||c}{45.2}           & \multicolumn{1}{c|}{29.1}             & 39.9          & 45.2           & 26.5             & 42.4             & 31.0             & 28.6             & \multicolumn{1}{c|}{17.1}             & \multicolumn{1}{c|}{29.1}             & 39.9          & 52.8                              & 37.8            & \multicolumn{1}{c|}{29.1}             & \multicolumn{1}{c|}{32.1}            & 39.9         \\
\hline
\end{tabular}
}
\end{table*}

\begin{table*}[h]
\centering
\captionsetup{font={small}}
    \caption{Ablation study on ADE20K dataset. Best in \textbf{bold}, runner-up {\ul underlined}.}
    \label{table:ablation}
    \vskip -0.3cm
\resizebox{\textwidth}{!}{
\begin{tabular}{lllccc||cccccccc||ccccc}
\hline
                                  &            &            & \multicolumn{3}{||c||}{\textbf{100-50}}                                   & \multicolumn{8}{c||}{\textbf{100-10}}                                                                                                                                                       & \multicolumn{5}{c}{\textbf{50-50}}                                                                                                                 \\\hline
\multicolumn{1}{l||}{\textbf{Method}} & \multicolumn{1}{l||}{\textbf{Architecture}} & \textbf{Backbone}   & \multicolumn{1}{||c}{\textsl{1-100}} & \multicolumn{1}{c|}{\textsl{101-150}} & \textsl{all}           & \textsl{1-100} & \textsl{101-110} & \textsl{111-120} & \textsl{121-130} & \textsl{131-140} & \multicolumn{1}{l|}{\textsl{141-150}} & \multicolumn{1}{l|}{\textsl{101-150}} & \textsl{all}  & \multicolumn{1}{l}{\textsl{1-50}} & \textsl{51-100} & \multicolumn{1}{c|}{\textsl{101-150}} & \multicolumn{1}{l|}{\textsl{51-150}} & \textsl{all}  \\ \hline
\multicolumn{1}{l||}{MiB}&   \multicolumn{1}{l||}{Transformer}          & ViT-small  & \multicolumn{1}{||c}{39.2}           & \multicolumn{1}{c|}{{28.7}}       & {35.7}    & 35.4           & {19.8}       & {26.4}       & {29.9}       & {14.8}       & \multicolumn{1}{c|}{15.3}             & \multicolumn{1}{c|}{{21.2}}       & 30.7          & 47.8                              & {33.9}      & \multicolumn{1}{c|}{{28.1}}       & \multicolumn{1}{c|}{{31.0}}      & {36.6}          \\
\multicolumn{1}{l||}{+$\mathcal{L}^t_{1}$}&   \multicolumn{1}{l||}{Transformer}          & ViT-small  & \multicolumn{1}{||c}{37.2}           & \multicolumn{1}{c|}{23.4}       & {32.6} & 32.7           & 17.1       & 24.4       & 24.9       & 12.2       & \multicolumn{1}{c|}{13.6}             & \multicolumn{1}{c|}{18.4}       & {28.0} & 44.8                              & 30.9      & \multicolumn{1}{c|}{25.4}       & \multicolumn{1}{c|}{28.2}      & {33.7}         \\
\multicolumn{1}{l||}{+$\mathcal{L}^t_{2}$}&   \multicolumn{1}{l||}{Transformer}          & ViT-small  & \multicolumn{1}{||c}{37.8}           & \multicolumn{1}{c|}{24.1}       & {33.2} & 33.2           & 17.8       & 24.2       & 26.8       & 11.4       & \multicolumn{1}{c|}{13.9}             & \multicolumn{1}{c|}{18.8}       & {28.4} & 45.2                              & 30.1      & \multicolumn{1}{c|}{25.2}       & \multicolumn{1}{c|}{27.7}      & {33.5}          \\
\multicolumn{1}{l||}{+$\mathcal{L}^t_{\text {cd}}$}&   \multicolumn{1}{l||}{Transformer}          & ViT-small  & \multicolumn{1}{||c}{\textbf{43.9}}           & \multicolumn{1}{c|}{26.0}       & {\ul 37.9} & \textbf{42.1}           & 18.9       & 25.8       & 29.3       & 13.8       & \multicolumn{1}{c|}{14.7}             & \multicolumn{1}{c|}{20.5}       & {\ul 34.9} & \textbf{50.8}                              & {32.3}      & \multicolumn{1}{c|}{{26.9}}       & \multicolumn{1}{c|}{{29.6}}      & {36.7}         \\
\multicolumn{1}{l||}{+$\mathcal{L}^t_{\text {ct}}$}&   \multicolumn{1}{l||}{Transformer}          & ViT-small  & \multicolumn{1}{||c}{38.7}           & \multicolumn{1}{c|}{\textbf{32.0}}       & 36.5 & 34.2           & \textbf{23.3}       & \textbf{31.5}       & \textbf{30.8}       & \textbf{15.9}       & \multicolumn{1}{c|}{\textbf{21.2}}             & \multicolumn{1}{c|}{\textbf{24.5}}       & {31.0} & 46.8                              & \textbf{}{35.8}      & \multicolumn{1}{c|}{\textbf{31.6}}       & \multicolumn{1}{c|}{\textbf{33.7}}      & {\ul 38.1}         \\
\multicolumn{1}{l||}{+$\mathcal{L}^t_{\text {cd}}$, $\mathcal{L}^t_{\text {ct}}$}&   \multicolumn{1}{l||}{Transformer}   & ViT-small   & \multicolumn{1}{||c}{{\ul43.0}}  & \multicolumn{1}{c|}{{\ul29.4}}    & \textbf{38.4} & {\ul41.2}  & {\ul21.1}    & {\ul29.3}    & {\ul30.6}    & {\ul15.0}    & \multicolumn{1}{c|}{{\ul19.8}}    & \multicolumn{1}{c|}{{\ul23.2}}    & \textbf{35.2} & {\ul49.7}                     & {\ul35.1}   & \multicolumn{1}{c|}{{\ul30.2}}    & \multicolumn{1}{c|}{{\ul32.7}}   & \textbf{38.3} \\ \hline
\end{tabular}
}
\vskip -0.5cm
\end{table*}

\subsection{ADE20K}
\label{sec:ade}

Differently from Pascal-VOC 2012, ADE20K contains both stuff(e.g. sky, building, wall) and object classes so that it is a more challenging semantic segmentation dataset. We follow \cite{cermelli2020modeling} to split the dataset into disjoint image sets without any constraint except ensuring a minimum number of images(i.e. 50) where classes on $\mathcal{C}_t$ have labeled pixels. Obviously, each $\mathcal{T}_t$ provides annotations only for classes in $\mathcal{C}_t$ while other classes(old or future) appear as background in the ground truth. We conduct three different experiments considering the addition of the last 50 classes at once(\textbf{100-50}), the addition of the last 50 classes 10 at a time(\textbf{100-10}) and the addition of the last 100 classes in 2 steps of 50 classes each(\textbf{50-50}). In the comparative experiments in Table \ref{table:ade}, we report the performance of different methods.  

We conduct the single-step addition of 50 classes(\textbf{100-50}) at first. As reported in Table \ref{table:ade}, all \textsl{FT} methods are clearly bad since they nearly forget all old knowledge. Obviously, $MiB_{T}$ suffers from a drop of $13.3\%$ in mIoU of the first 100 classes compared with offline training, which is lower than the drop in mIoU of the first 100 classes of $MiB_{D}$($14.4\%$). This is a further proof that Transformer based architecture is more applicable to suppress catastrophic forgetting problems on ISS. Meanwhile, \textsl{TISS} surpasses over the best baseline of $2.6\%$ on old classes, of $2.4\%$ on new classes and $7.6\%$ on all classes.

\begin{figure}[h]
\centering
\includegraphics[width=\linewidth]{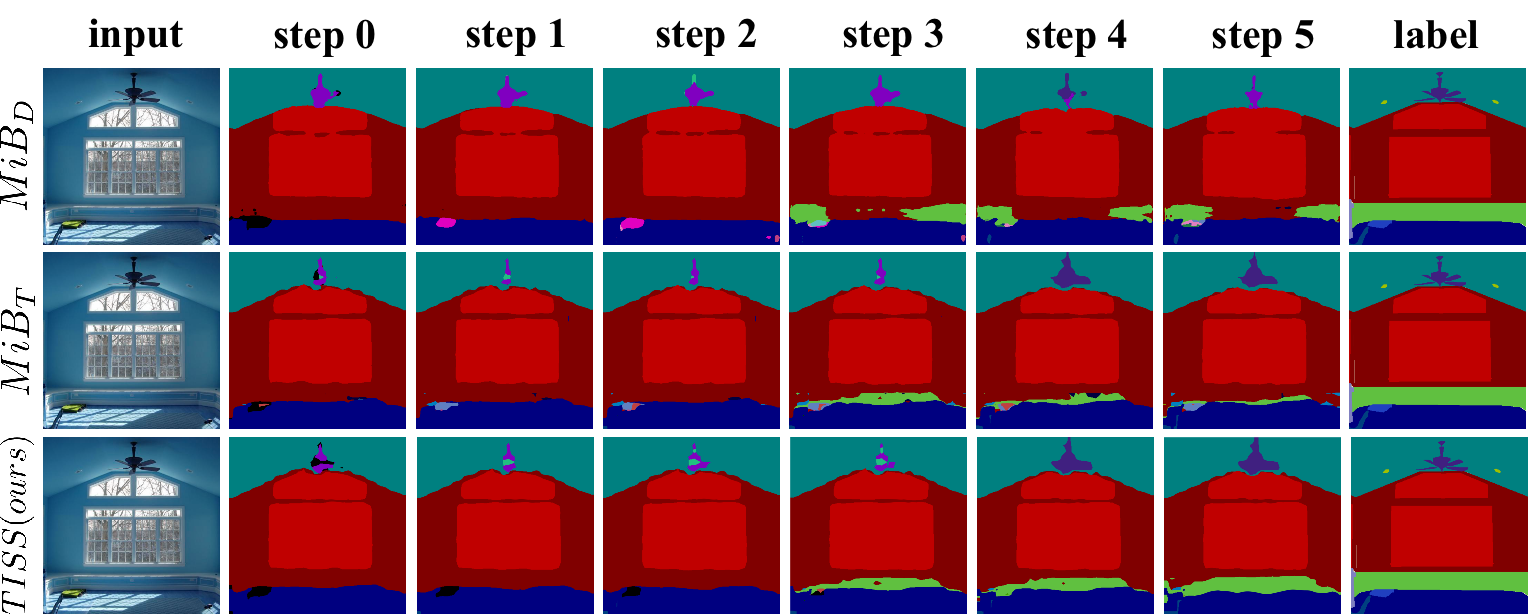}
\caption{Qualitative results at each step of ADE20K \textbf{100-10} task using three methods: $MiB_D$, $MiB_T$ and $TISS(ours)$.} 
\label{fig:vis}
\vskip -0.5cm
\end{figure}

Then we carry out the multi-step addition of 50 classes(\textbf{100-10}). All results are reported in Table \ref{table:ade}. Obviously, the catastrophic forgetting problems on CNNs based architecture are more severe that $MiB_{D}$ suffers from drop of $40.2\%$ in mIoU of the first 100 classes compared with the corresponding offline learning, while the drop in mIoU of the first 100 classes between $MiB_{T}$ and the corresponding offline learning is only $21.7\%$. Furthermore, \textsl{TISS} achieves boosts over the best baseline of $1.7\%$ on old classes, of $17.4\%$ on new classes and $9.2\%$ on all classes. In Figure \ref{fig:vis} we present some qualitative results which demonstrate the superiority of our method. Visualization results on other tasks can be found in appendix.

Finally, in Table \ref{table:ade} we analyze the performance on three sequential steps of 50 classes(\textbf{50-50}). $MiB_{T}$ suffers from decrease of $9.5\%$ in mIoU of the first 50 classes compared with the corresponding offline learning, while the drop in mIoU of the first 50 classes of $MiB_{D}$ is $26.9\%$. At the same time, \textsl{TISS} achieves improvements over the best baseline of $4.1\%$ on old classes, of $7.5\%$ on new classes and $6.3\%$ on all classes.

In addition, as shown in Table \ref{table:voc} and Table \ref{table:ade}, we find that freezing BN layers after the first step is effective in suppressing catastrophic forgetting($4.8\%$ average improvement on old classes for Pascal-VOC 2012 and $3.5\%$ average improvement on old classes ADE20K), which further corroborates that the extensive application of Batch Normalization(BN) layers in CNNs leads to catastrophic forgetting.

\subsection{Ablation Study}

\vskip -0.1cm

In Table \ref{table:ablation} we report a detailed analysis of our contributions. Considering three ADE20K tasks, we start from the baseline \textsl{MiB} method applying Transform based architecture. We first add $\mathcal{L}_1$ distillation loss $\mathcal{L}^t_1$ and $\mathcal{L}_2$ distillation loss $\mathcal{L}^t_2$ respectively, which degrade the performance on all metrics. Then we only add the patch-wise contrastive distillation loss $\mathcal{L}^t_{\text {cd}}$ to the baseline, which improves the capability of Transform based architecture to accumulate knowledge($12.4\%$ average improvement on old classes). Second, we only add the patch-wise contrastive loss $\mathcal{L}^t_{\text {ct}}$ to the baseline, enhancing the generalization capability of Transformer based architecture on new tasks($11.9\%$ average improvement on new classes). Finally, we add both $\mathcal{L}^t_{\text {cd}}$ and $\mathcal{L}^t_{\text {ct}}$ to the baseline, which provides boosts on the performances for both old and new classes($10.0\%$ average improvement on old classes, $5.8\%$ average improvement on new classes and $9.0\%$ average improvement on all classes), which shows that the two losses provide mutual benefits.

\section{Conclusions}
In this work we first investigate the adaptation problem of different architectures on incremental semantic segmentation and analyze the bottlenecks of CNNs based architectures on ISS. Furthermore, we innovatively introduce \textsl{TISS} which combines the distillation based method with Transformer based architecture, effectively learning new classes without weakening its ability to recognize old ones. In addition, we improve the feature diversification of Transformer by adding two patch-wise contrastive losses, resulting in substantial improvements. Experiments show that our method significantly outperforms many state-of-the-art methods in both Pascal-VOC 2012 and ADE20K datasets. We hope that our work can shed new light on the development of the field of incremental semantic segmentation.

{\small
\bibliographystyle{ieee_fullname}
\bibliography{egbib}

\begin{thebibliography}{10}\itemsep=-1pt

\bibitem{ba2016layer}
Jimmy~Lei Ba, Jamie~Ryan Kiros, and Geoffrey~E Hinton.
\newblock Layer normalization.
\newblock {\em arXiv preprint arXiv:1607.06450}, 2016.

\bibitem{bao2021beit}
Hangbo Bao, Li Dong, and Furu Wei.
\newblock Beit: Bert pre-training of image transformers.
\newblock {\em arXiv preprint arXiv:2106.08254}, 2021.

\bibitem{carion2020end}
Nicolas Carion, Francisco Massa, Gabriel Synnaeve, Nicolas Usunier, Alexander
  Kirillov, and Sergey Zagoruyko.
\newblock End-to-end object detection with transformers.
\newblock In {\em European conference on computer vision}, pages 213--229.
  Springer, 2020.

\bibitem{cermelli2021incremental}
Fabio Cermelli, Dario Fontanel, Antonio Tavera, Marco Ciccone, and Barbara
  Caputo.
\newblock Incremental learning in semantic segmentation from image labels.
\newblock {\em arXiv preprint arXiv:2112.01882}, 2021.

\bibitem{cermelli2020modeling}
Fabio Cermelli, Massimiliano Mancini, Samuel~Rota Bulo, Elisa Ricci, and
  Barbara Caputo.
\newblock Modeling the background for incremental learning in semantic
  segmentation.
\newblock In {\em Proceedings of the IEEE/CVF Conference on Computer Vision and
  Pattern Recognition}, pages 9233--9242, 2020.

\bibitem{chen2017rethinking}
Liang-Chieh Chen, George Papandreou, Florian Schroff, and Hartwig Adam.
\newblock Rethinking atrous convolution for semantic image segmentation.
\newblock {\em arXiv preprint arXiv:1706.05587}, 2017.

\bibitem{chen2018encoder}
Liang-Chieh Chen, Yukun Zhu, George Papandreou, Florian Schroff, and Hartwig
  Adam.
\newblock Encoder-decoder with atrous separable convolution for semantic image
  segmentation.
\newblock In {\em Proceedings of the European conference on computer vision
  (ECCV)}, pages 801--818, 2018.

\bibitem{cheng2021masked}
Bowen Cheng, Ishan Misra, Alexander~G Schwing, Alexander Kirillov, and Rohit
  Girdhar.
\newblock Masked-attention mask transformer for universal image segmentation.
\newblock {\em arXiv preprint arXiv:2112.01527}, 2021.

\bibitem{cheng2021per}
Bowen Cheng, Alex Schwing, and Alexander Kirillov.
\newblock Per-pixel classification is not all you need for semantic
  segmentation.
\newblock {\em Advances in Neural Information Processing Systems}, 34, 2021.

\bibitem{chu2021conditional}
Xiangxiang Chu, Zhi Tian, Bo Zhang, Xinlong Wang, Xiaolin Wei, Huaxia Xia, and
  Chunhua Shen.
\newblock Conditional positional encodings for vision transformers.
\newblock {\em arXiv preprint arXiv:2102.10882}, 2021.

\bibitem{cordts2016cityscapes}
Marius Cordts, Mohamed Omran, Sebastian Ramos, Timo Rehfeld, Markus Enzweiler,
  Rodrigo Benenson, Uwe Franke, Stefan Roth, and Bernt Schiele.
\newblock The cityscapes dataset for semantic urban scene understanding.
\newblock In {\em Proceedings of the IEEE conference on computer vision and
  pattern recognition}, pages 3213--3223, 2016.

\bibitem{de2019continual}
Matthias De~Lange, Rahaf Aljundi, Marc Masana, Sarah Parisot, Xu Jia, Ales
  Leonardis, Gregory Slabaugh, and Tinne Tuytelaars.
\newblock Continual learning: A comparative study on how to defy forgetting in
  classification tasks.
\newblock {\em arXiv preprint arXiv:1909.08383}, 2(6), 2019.

\bibitem{dosovitskiy2020image}
Alexey Dosovitskiy, Lucas Beyer, Alexander Kolesnikov, Dirk Weissenborn,
  Xiaohua Zhai, Thomas Unterthiner, Mostafa Dehghani, Matthias Minderer, Georg
  Heigold, Sylvain Gelly, et~al.
\newblock An image is worth 16x16 words: Transformers for image recognition at
  scale.
\newblock {\em arXiv preprint arXiv:2010.11929}, 2020.

\bibitem{douillard2021tackling}
Arthur Douillard, Yifu Chen, Arnaud Dapogny, and Matthieu Cord.
\newblock Tackling catastrophic forgetting and background shift in continual
  semantic segmentation.
\newblock {\em arXiv preprint arXiv:2106.15287}, 2021.

\bibitem{pascal-voc-2012}
M. Everingham, L. Van~Gool, C.~K.~I. Williams, J. Winn, and A. Zisserman.
\newblock The {PASCAL} {V}isual {O}bject {C}lasses {C}hallenge 2012 {(VOC2012)}
  {R}esults.
\newblock
  http://www.pascal-network.org/challenges/VOC/voc2012/workshop/index.html.

\bibitem{french1999catastrophic}
Robert~M French.
\newblock Catastrophic forgetting in connectionist networks.
\newblock {\em Trends in cognitive sciences}, 3(4):128--135, 1999.

\bibitem{gong2021vision}
Chengyue Gong, Dilin Wang, Meng Li, Vikas Chandra, and Qiang Liu.
\newblock Vision transformers with patch diversification.
\newblock {\em arXiv preprint arXiv:2104.12753}, 2021.

\bibitem{goodfellow2013empirical}
Ian~J Goodfellow, Mehdi Mirza, Da Xiao, Aaron Courville, and Yoshua Bengio.
\newblock An empirical investigation of catastrophic forgetting in
  gradient-based neural networks.
\newblock {\em arXiv preprint arXiv:1312.6211}, 2013.

\bibitem{han2021transformer}
Kai Han, An Xiao, Enhua Wu, Jianyuan Guo, Chunjing Xu, and Yunhe Wang.
\newblock Transformer in transformer.
\newblock {\em Advances in Neural Information Processing Systems}, 34, 2021.

\bibitem{he2016deep}
Kaiming He, Xiangyu Zhang, Shaoqing Ren, and Jian Sun.
\newblock Deep residual learning for image recognition.
\newblock In {\em Proceedings of the IEEE conference on computer vision and
  pattern recognition}, pages 770--778, 2016.

\bibitem{ioffe2015batch}
Sergey Ioffe and Christian Szegedy.
\newblock Batch normalization: Accelerating deep network training by reducing
  internal covariate shift.
\newblock In {\em International conference on machine learning}, pages
  448--456. PMLR, 2015.

\bibitem{jain2021semask}
Jitesh Jain, Anukriti Singh, Nikita Orlov, Zilong Huang, Jiachen Li, Steven
  Walton, and Humphrey Shi.
\newblock Semask: Semantically masked transformers for semantic segmentation.
\newblock {\em arXiv preprint arXiv:2112.12782}, 2021.

\bibitem{jiang2021token}
Zihang Jiang, Qibin Hou, Li Yuan, Daquan Zhou, Xiaojie Jin, Anran Wang, and
  Jiashi Feng.
\newblock Token labeling: Training a 85.4\% top-1 accuracy vision transformer
  with 56m parameters on imagenet.
\newblock {\em arXiv e-prints}, pages arXiv--2104, 2021.

\bibitem{kirkpatrick2017overcoming}
James Kirkpatrick, Razvan Pascanu, Neil Rabinowitz, Joel Veness, Guillaume
  Desjardins, Andrei~A Rusu, Kieran Milan, John Quan, Tiago Ramalho, Agnieszka
  Grabska-Barwinska, et~al.
\newblock Overcoming catastrophic forgetting in neural networks.
\newblock {\em Proceedings of the national academy of sciences},
  114(13):3521--3526, 2017.

\bibitem{klingner2020class}
Marvin Klingner, Andreas B{\"a}r, Philipp Donn, and Tim Fingscheidt.
\newblock Class-incremental learning for semantic segmentation re-using neither
  old data nor old labels.
\newblock In {\em 2020 IEEE 23rd International Conference on Intelligent
  Transportation Systems (ITSC)}, pages 1--8. IEEE, 2020.

\bibitem{li2022technical}
Duo Li, Guimei Cao, Yunlu Xu, Zhanzhan Cheng, and Yi Niu.
\newblock Technical report for iccv 2021 challenge sslad-track3b: Transformers
  are better continual learners.
\newblock {\em arXiv preprint arXiv:2201.04924}, 2022.

\bibitem{li2019rilod}
Dawei Li, Serafettin Tasci, Shalini Ghosh, Jingwen Zhu, Junting Zhang, and
  Larry Heck.
\newblock Rilod: Near real-time incremental learning for object detection at
  the edge.
\newblock In {\em Proceedings of the 4th ACM/IEEE Symposium on Edge Computing},
  pages 113--126, 2019.

\bibitem{li2017learning}
Zhizhong Li and Derek Hoiem.
\newblock Learning without forgetting.
\newblock {\em IEEE transactions on pattern analysis and machine intelligence},
  40(12):2935--2947, 2017.

\bibitem{lin2017refinenet}
Guosheng Lin, Anton Milan, Chunhua Shen, and Ian Reid.
\newblock Refinenet: Multi-path refinement networks for high-resolution
  semantic segmentation.
\newblock In {\em Proceedings of the IEEE conference on computer vision and
  pattern recognition}, pages 1925--1934, 2017.

\bibitem{liu2021swin}
Ze Liu, Yutong Lin, Yue Cao, Han Hu, Yixuan Wei, Zheng Zhang, Stephen Lin, and
  Baining Guo.
\newblock Swin transformer: Hierarchical vision transformer using shifted
  windows.
\newblock In {\em Proceedings of the IEEE/CVF International Conference on
  Computer Vision}, pages 10012--10022, 2021.

\bibitem{long2015fully}
Jonathan Long, Evan Shelhamer, and Trevor Darrell.
\newblock Fully convolutional networks for semantic segmentation.
\newblock In {\em Proceedings of the IEEE conference on computer vision and
  pattern recognition}, pages 3431--3440, 2015.

\bibitem{maracani2021recall}
Andrea Maracani, Umberto Michieli, Marco Toldo, and Pietro Zanuttigh.
\newblock Recall: Replay-based continual learning in semantic segmentation.
\newblock In {\em Proceedings of the IEEE/CVF International Conference on
  Computer Vision}, pages 7026--7035, 2021.

\bibitem{mccloskey1989catastrophic}
Michael McCloskey and Neal~J Cohen.
\newblock Catastrophic interference in connectionist networks: The sequential
  learning problem.
\newblock In {\em Psychology of learning and motivation}, volume~24, pages
  109--165. Elsevier, 1989.

\bibitem{michieli2019incremental}
Umberto Michieli and Pietro Zanuttigh.
\newblock Incremental learning techniques for semantic segmentation.
\newblock In {\em Proceedings of the IEEE/CVF International Conference on
  Computer Vision Workshops}, pages 0--0, 2019.

\bibitem{michieli2021continual}
Umberto Michieli and Pietro Zanuttigh.
\newblock Continual semantic segmentation via repulsion-attraction of sparse
  and disentangled latent representations.
\newblock In {\em Proceedings of the IEEE/CVF Conference on Computer Vision and
  Pattern Recognition}, pages 1114--1124, 2021.

\bibitem{michieli2021knowledge}
Umberto Michieli and Pietro Zanuttigh.
\newblock Knowledge distillation for incremental learning in semantic
  segmentation.
\newblock {\em Computer Vision and Image Understanding}, 2021.

\bibitem{qu2021recent}
Haoxuan Qu, Hossein Rahmani, Li Xu, Bryan Williams, and Jun Liu.
\newblock Recent advances of continual learning in computer vision: An
  overview.
\newblock {\em arXiv preprint arXiv:2109.11369}, 2021.

\bibitem{raghu2021vision}
Maithra Raghu, Thomas Unterthiner, Simon Kornblith, Chiyuan Zhang, and Alexey
  Dosovitskiy.
\newblock Do vision transformers see like convolutional neural networks?
\newblock {\em Advances in Neural Information Processing Systems}, 34, 2021.

\bibitem{shmelkov2017incremental}
Konstantin Shmelkov, Cordelia Schmid, and Karteek Alahari.
\newblock Incremental learning of object detectors without catastrophic
  forgetting.
\newblock In {\em Proceedings of the IEEE international conference on computer
  vision}, pages 3400--3409, 2017.

\bibitem{strudel2021segmenter}
Robin Strudel, Ricardo Garcia, Ivan Laptev, and Cordelia Schmid.
\newblock Segmenter: Transformer for semantic segmentation.
\newblock In {\em Proceedings of the IEEE/CVF International Conference on
  Computer Vision}, pages 7262--7272, 2021.

\bibitem{tasar2019incremental}
Onur Tasar, Yuliya Tarabalka, and Pierre Alliez.
\newblock Incremental learning for semantic segmentation of large-scale remote
  sensing data.
\newblock {\em IEEE Journal of Selected Topics in Applied Earth Observations
  and Remote Sensing}, 12(9):3524--3537, 2019.

\bibitem{touvron2021training}
Hugo Touvron, Matthieu Cord, Matthijs Douze, Francisco Massa, Alexandre
  Sablayrolles, and Herv{\'e} J{\'e}gou.
\newblock Training data-efficient image transformers \& distillation through
  attention.
\newblock In {\em International Conference on Machine Learning}, pages
  10347--10357. PMLR, 2021.

\bibitem{touvron2021going}
Hugo Touvron, Matthieu Cord, Alexandre Sablayrolles, Gabriel Synnaeve, and
  Herv{\'e} J{\'e}gou.
\newblock Going deeper with image transformers.
\newblock In {\em Proceedings of the IEEE/CVF International Conference on
  Computer Vision}, pages 32--42, 2021.

\bibitem{vaswani2017attention}
Ashish Vaswani, Noam Shazeer, Niki Parmar, Jakob Uszkoreit, Llion Jones,
  Aidan~N Gomez, {\L}ukasz Kaiser, and Illia Polosukhin.
\newblock Attention is all you need.
\newblock {\em Advances in neural information processing systems}, 30, 2017.

\bibitem{xie2021segformer}
Enze Xie, Wenhai Wang, Zhiding Yu, Anima Anandkumar, Jose~M Alvarez, and Ping
  Luo.
\newblock Segformer: Simple and efficient design for semantic segmentation with
  transformers.
\newblock {\em Advances in Neural Information Processing Systems}, 34, 2021.

\bibitem{yan2021framework}
Shipeng Yan, Jiale Zhou, Jiangwei Xie, Songyang Zhang, and Xuming He.
\newblock An em framework for online incremental learning of semantic
  segmentation.
\newblock In {\em Proceedings of the 29th ACM International Conference on
  Multimedia}, pages 3052--3060, 2021.

\bibitem{yang2022continual}
Guanglei Yang, Enrico Fini, Dan Xu, Paolo Rota, Mingli Ding, Hao Tang, Xavier
  Alameda-Pineda, and Elisa Ricci.
\newblock Continual attentive fusion for incremental learning in semantic
  segmentation.
\newblock {\em arXiv preprint arXiv:2202.00432}, 2022.

\bibitem{zhang2018exfuse}
Zhenli Zhang, Xiangyu Zhang, Chao Peng, Xiangyang Xue, and Jian Sun.
\newblock Exfuse: Enhancing feature fusion for semantic segmentation.
\newblock In {\em Proceedings of the European conference on computer vision
  (ECCV)}, pages 269--284, 2018.

\bibitem{zhao2017pyramid}
Hengshuang Zhao, Jianping Shi, Xiaojuan Qi, Xiaogang Wang, and Jiaya Jia.
\newblock Pyramid scene parsing network.
\newblock In {\em Proceedings of the IEEE conference on computer vision and
  pattern recognition}, pages 2881--2890, 2017.

\bibitem{zhou2017scene}
Bolei Zhou, Hang Zhao, Xavier Puig, Sanja Fidler, Adela Barriuso, and Antonio
  Torralba.
\newblock Scene parsing through ade20k dataset.
\newblock In {\em Proceedings of the IEEE conference on computer vision and
  pattern recognition}, pages 633--641, 2017.

\end{thebibliography}
}

\newpage
\section{Appendix}
\subsection{Experiments on Cityscapes\cite{cordts2016cityscapes}}
\vskip -0.5cm
\begin{table}[h]
\centering
    \caption{Comparison of Cityscapes \textbf{14-1} task between different methods. Best in \textbf{bold}, runner-up {\ul underlined}, $\star$: results from [13].}
    \label{table:cityscapes}
\vskip -0.3cm
\resizebox{0.5\textwidth}{!}{
\begin{tabular}{lll||cccccccc}
\hline
\multicolumn{1}{l||}{\textbf{Method}} & \multicolumn{1}{l||}{\textbf{Architecture}} & \textbf{Backbone} & \textsl{1-14} & \textsl{15} & \textsl{16} & \textsl{17} & \textsl{18} & \multicolumn{1}{l|}{\textsl{19}} & \multicolumn{1}{l|}{\textsl{15-19}} & \textsl{all}\\ \hline
\multicolumn{1}{l||}{MiB$\star$}&   \multicolumn{1}{l||}{Deeplab-v3}          & ResNet-101  & 55.1          & -    & -      & -      & -    & \multicolumn{1}{c|}{-}             & \multicolumn{1}{c|}{12.9}       &44.6        \\ 
\multicolumn{1}{l||}{PLOP$\star$}&   \multicolumn{1}{l||}{Deeplab-v3}          & ResNet-101  & 56.6          & -    & -      & -      & -    & \multicolumn{1}{c|}{-}             & \multicolumn{1}{c|}{13.1}       &45.7       \\
\multicolumn{1}{l||}{PLOPLong$\star$}&   \multicolumn{1}{l||}{Deeplab-v3}          & ResNet-101  & 57.8          & -    & -      & -      & -    & \multicolumn{1}{c|}{-}             & \multicolumn{1}{c|}{23.1}       &49.2   \\   
\multicolumn{1}{l||}{MiB}&   \multicolumn{1}{l||}{Transformer}          & ViT-base  & {\ul 59.5}          & \textbf{62.3} & {\ul 58.7} & {\ul 36.1} & {\ul 34.1} & \multicolumn{1}{c|}{{\ul 48.5}}             & \multicolumn{1}{c|}{{\ul 47.9}}       & {\ul 56.5}   \\
\multicolumn{1}{l||}{TISS}&   \multicolumn{1}{l||}{Transformer}          & ViT-base   & \textbf{61.9}          & {\ul 57.2}  & \textbf{61.7} & \textbf{39.2} & \textbf{35.1} & \multicolumn{1}{c|}{\textbf{49.4}}             & \multicolumn{1}{c|}{\textbf{48.5}}       &\textbf{58.4}        \\ \hline
\end{tabular}
}
\vskip -0.3cm
\end{table}

In addition to Pascal-VOC 2012\cite{pascal-voc-2012} and ADE20K\cite{zhou2017scene}, we also conduct \textbf{14-1} task(adding the remaining 5 classes sequentially) on Cityscapes\cite{cordts2016cityscapes} as shown in Table \ref{table:cityscapes}, for \textbf{14-1} task, \textsl{TISS} surpasses \textsl{MiB} with CNN architecture by $12.3\%$ in the first 14 classes, $256.0\%$ in the added 5 classes and $30.9\%$ in all classes. When it comes to \textsl{PLOPLong}, \textsl{TISS} achieves an average improvement over \textsl{PLOPLong} of $7.1\%$ in the first 14 classes, $110.0\%$ in the added 5 classes and $18.7\%$ in all classes. Finally, \textsl{TISS} surpasses over \textsl{MiB} with Transformer architecture of $4.0\%$ in the first 14 classes, $1.3\%$ in the added 5 classes and $3.4\%$ in all classes, which reinforces the superiority of \textsl{TISS}.

\subsection{Experiments on more Transformer models}
\vskip -0.5cm
\begin{table}[h]
\centering
    \caption{ADE20K \textbf{100-10} task with SegFormer\cite{xie2021segformer} and Swin-Transformer\cite{liu2021swin}. Best in \textbf{bold}, runner-up {\ul underlined}.}
    \label{table:swin}
\vskip -0.3cm
\resizebox{0.5\textwidth}{!}{
\begin{tabular}{ll||cccccccc}
\hline
\multicolumn{1}{l||}{\textbf{Method}} & \multicolumn{1}{l||}{\textbf{Architecture}} & \textsl{1-100} & \textsl{101-110} & \textsl{111-120} & \textsl{121-130} & \textsl{131-140} & \multicolumn{1}{l|}{\textsl{141-150}} & \multicolumn{1}{l|}{\textsl{101-150}} & \textsl{all}\\\hline

\multicolumn{1}{l||}{MiB}&   \multicolumn{1}{l||}{Seg-mit3} & 40.3           & 20.4       & {\ul 42.9}       & {\ul 40.2}       & 27.0      & \multicolumn{1}{c|}{23.0}             & \multicolumn{1}{c|}{30.7}       & 37.1         \\

\multicolumn{1}{l||}{MiB}&   \multicolumn{1}{l||}{Swin-base}& { 41.7}           & {\ul 27.1}     & 39.9       & {38.2}       & {\ul 30.3}      & \multicolumn{1}{c|}{{25.4}}             & \multicolumn{1}{c|}{{32.1}}       & {38.2}         \\

\multicolumn{1}{l||}{TISS(ours)}&   \multicolumn{1}{l||}{Seg-mit3} & {\ul 42.1}  & 21.8    & \textbf{ 43.7}    & { 39.7}    & { 29.8}    & \multicolumn{1}{c|}{{\ul 25.7}}    & \multicolumn{1}{c|}{{\ul 32.2}}    & {\ul 38.8} \\

\multicolumn{1}{l||}{TISS(ours)}&   \multicolumn{1}{l||}{Swin-base} & \textbf{ 43.1}  & \textbf{ 29.2}    & 38.0    & \textbf{ 41.1}    & \textbf{ 32.4}    & \multicolumn{1}{c|}{\textbf{ 26.2}}    & \multicolumn{1}{c|}{\textbf{ 33.4}}    & \textbf{39.9} \\ \hline
\end{tabular}
}
\vskip -0.3cm
\end{table}
The model in our work is based on Segmenter\cite{strudel2021segmenter} with ViT backbone. We also conduct ADE20K 100-10 task with Swin-Transformer\cite{liu2021swin} and Segformer\cite{xie2021segformer} as shown in Table \ref{table:swin}, when it comes to Swin-Transformer, \textsl{TISS} surpasses \textsl{MiB} by $3.4\%$ in the first 100 classes, $4.0\%$ in the added 50 classes and $4.5\%$ in all classes. For Segformer, \textsl{TISS} achieves an average improvement over \textsl{MiB} of $4.5\%$ in the first 100 classes, $4.9\%$ in the added 50 classes and $4.6\%$ in all classes. Consequently, our method is still effective on both Swin-Transformer and Segformer. 

\subsection{Visualization}

\begin{figure}[!htbp]
\vskip -0.3cm
\centering
\includegraphics[width=\linewidth]{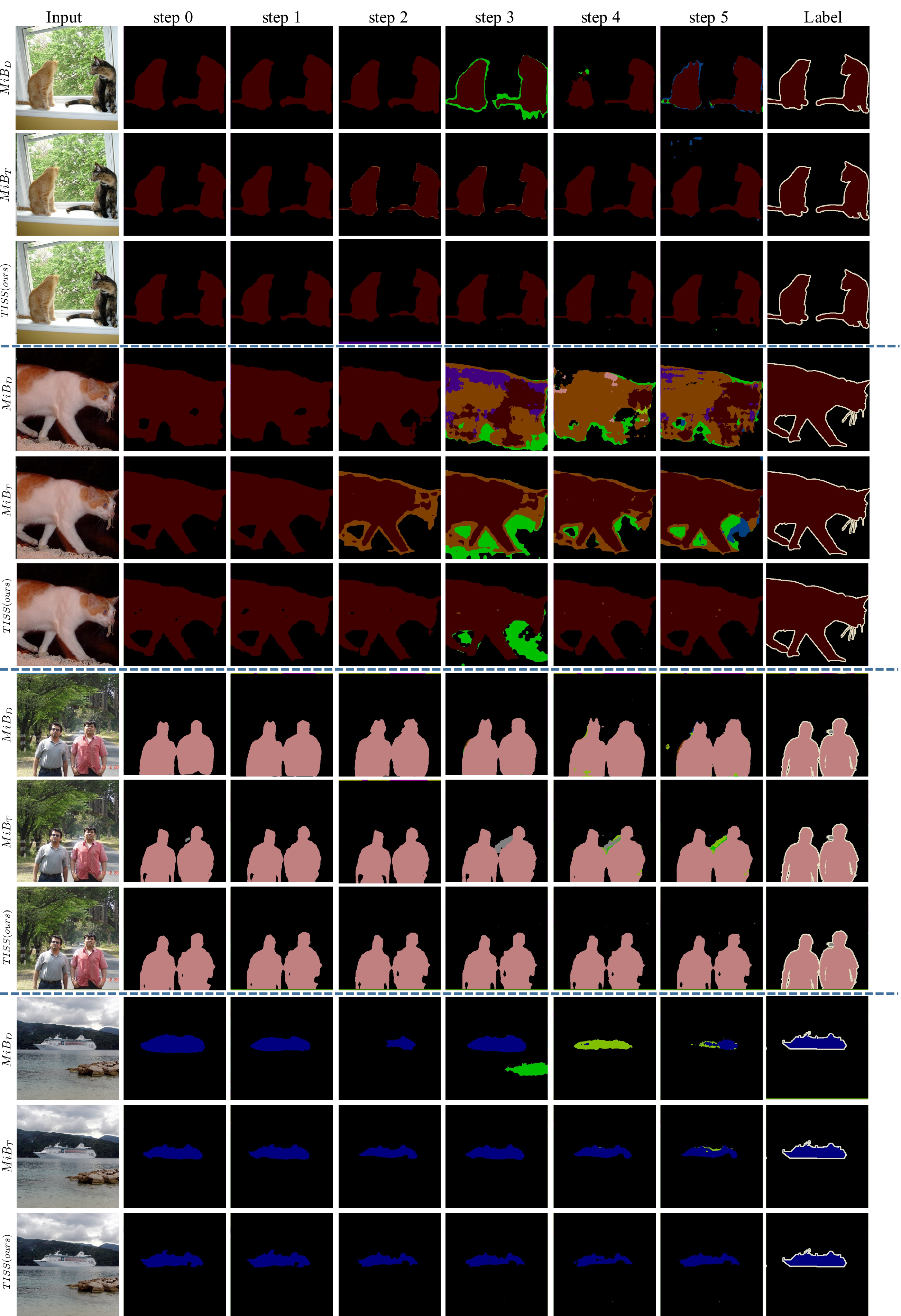}
\caption{Qualitative results at each step of Pascal-VOC 2012 \textbf{15-1} task using three methods: $MiB_D$, $MiB_T$ and $TISS(ours)$.} 
\label{fig:vis-voc}
\vskip -0.3cm
\end{figure}

\begin{figure}[h]

\centering
\includegraphics[width=\linewidth]{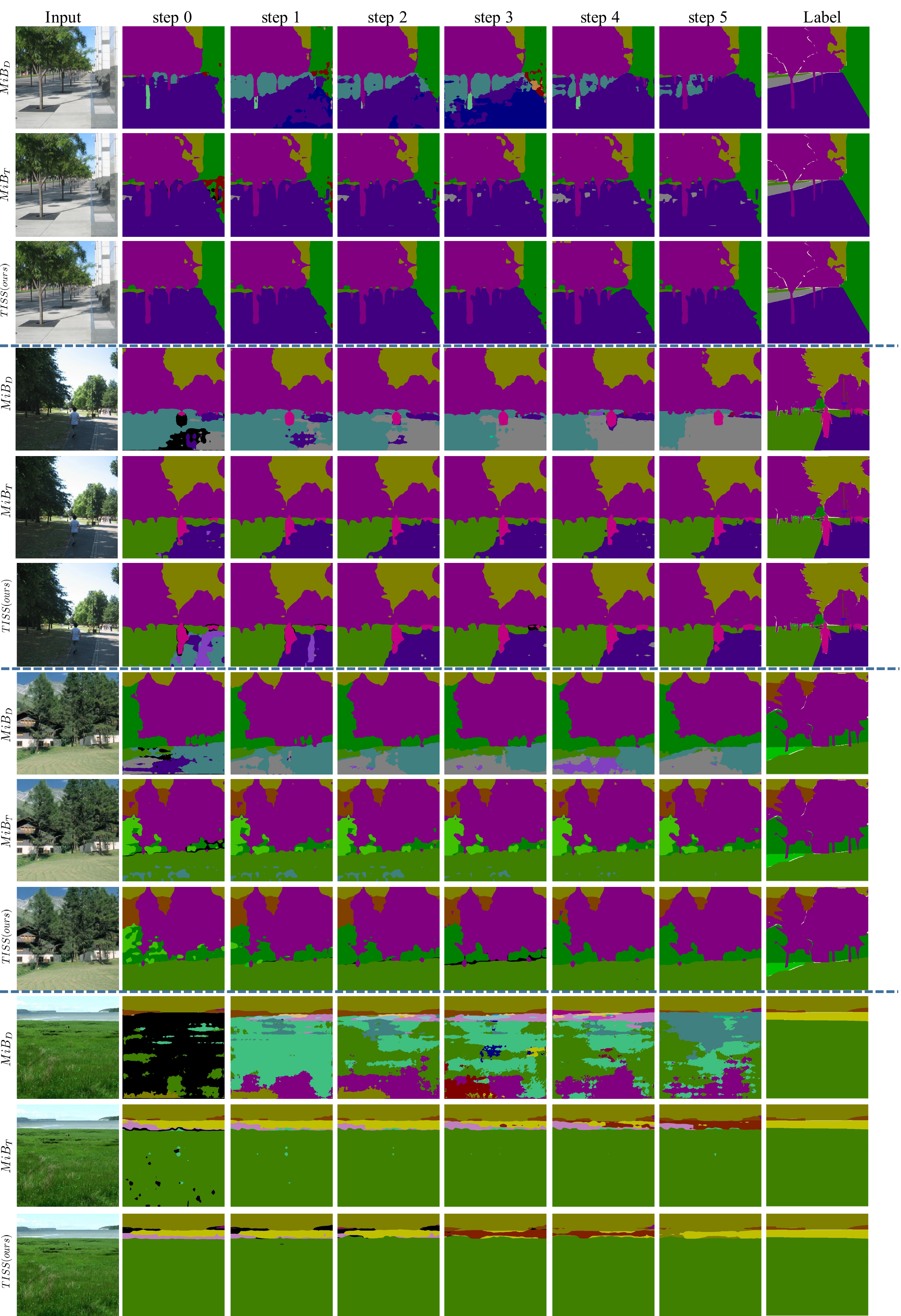}
\caption{Qualitative results at each step of ADE20K \textbf{100-10} task using three methods: $MiB_D$, $MiB_T$ and $TISS(ours)$.} 
\label{fig:vis-ade}

\end{figure}

As Figure \ref{fig:vis-voc} and Figure \ref{fig:vis-ade} show, we present qualitative results on Pascal-VOC 2012 \textbf{15-1} task and ADE20K \textbf{100-10} task, which intuitively demonstrate the superiority of \textsl{TISS}.

\subsection{Per Class Results on Pascal-VOC 2012}
From Table \ref{table:19-1d} to \ref{table:15-1o}, we report the results for all classes of the Pascal-VOC 2012 dataset. We compare several methods such as CNNs based method \textsl{MiB}\cite{cermelli2020modeling}(we report both the results from \cite{cermelli2020modeling} and the implementation by ourselves) and \textsl{ILT}\cite{michieli2019incremental}, Transformer based method \textsl{MiB}\cite{cermelli2020modeling} and \textsl{TISS}. As the tables show, in \textsl{MiB} method, although 
in some classes(\textsl{e.g.} bus and cow in \textbf{19-1} task), CNNs based architecture applying the Deeplab-v3 architecture with ResNet-101 as backbone(denoted as $MiB_{D}$) surpasses the Transformer based architecture with ViT-small as backbone(denoted as ${MiB}_{T}$), $MiB_{T}$ improves the overall performance on all tasks in both \textsl{disjoint} and \textsl{overlapped} scenarios, which corroborates the effectiveness of Transformer based architecture on Incremental Semantic Segmentation(ISS). Furthermore, \textsl{TISS} outperforms other methods in 
almost every class for all tasks in both \textsl{disjoint} and \textsl{overlapped} scenarios(except table in the \textsl{disjoint} scenario of \textbf{19-1} task), which emphasizes the capability of \textsl{TISS} to not only accumulate previous knowledge while learning new knowledge, but also to learn discriminative features for difficult cases during different learning steps.

\begin{table*}[!htbp]

\centering
    \caption{Per Class mIoU on \textbf{19-1} task of Pascal-VOC 2012 under \textsl{disjoint} setup. Best in \textbf{bold}, runner-up {\ul underlined}. $\dagger$: results from \cite{cermelli2020modeling}}
    \label{table:19-1d}
\vskip -0.3cm
\resizebox{\textwidth}{!}{
\begin{tabular}{lll||ccccccccccccccccccc|c||c|c}
\hline
\multicolumn{1}{l||}{\textbf{Method}} & \multicolumn{1}{l||}{\textbf{Architecture}} & \textbf{Backbone} & aero & bike & bird & boat & bottle & bus & car & cat & chair & cow & table & dog & horse & mbike & person & plant & sheep & sofa & \multicolumn{1}{c|}{train} & tv & \textbf{1-19} & \textbf{all}\\ \hline
\multicolumn{1}{l||}{FT$\dagger$} & \multicolumn{1}{l||}{Deeplab-v3} & ResNet-101 & 11.9 & 2.1 & 1.1 & 11.6 & 4.8 & 6.9 & 13.5 & 0.2 & 0.0 & 3.8 & 14.4 & 0.5 & 1.5 & 4.7 & 0.0 & 15.8 & 2.8 & 1.8 & \multicolumn{1}{c|}{13.5} & 12.3 & 5.8 & 6.2\\
\multicolumn{1}{l||}{FT} & \multicolumn{1}{l||}{Transformer} & ViT-small & 10.7 & 2.5 & 1.3 & 9.2 & 6.8 & 8.1 & 9.2 & 0.5 & 0.3 & 2.2 & 15.9 & 0.9 & 0.7 & 3.7 & 0.0 & 18.7 & 3.4 & 2.6 & \multicolumn{1}{c|}{14.7} & 50.9 & 5.9 & 8.1 \\ \hline
\multicolumn{1}{l||}{MiB$\dagger$} & \multicolumn{1}{l||}{Deeplab-v3} & ResNet-101 & 78.0 & 40.5 & 85.7 & 51.6 & 64.4 & {\ul 79.1} & 77.8 & {\ul 89.9} & 39.2 & {\ul 82.3} & 55.4 & {\ul 86.2} & 82.7 & 72.2 & 83.6 & 56.6 & {\ul 86.2} & 45.1 & \multicolumn{1}{c|}{{\ul 65.0}} & 25.6 & 69.6 & 67.4\\
\multicolumn{1}{l||}{ILT$\dagger$} & \multicolumn{1}{l||}{Deeplab-v3} & ResNet-101 & 83.7 & {\ul 40.8} & 80.8 & 59.1 & 58.4 & 77.6 & {\ul 82.4} & 82.3 & 38.9 & 81.7 & 50.8 & 84.8 & {\ul 86.6} & {\ul 81.0} & 83.3 & 56.4 & 82.2 & 43.8 & \multicolumn{1}{c|}{57.5} & 16.4 & 69.1 & 66.4\\
\multicolumn{1}{l||}{MiB} & \multicolumn{1}{l||}{Deeplab-v3} & ResNet-101 & 82.5 & 38.9 & 85.7 & 56.9 & 60.4 & 41.7 & 79.7 & 89.1 & 32.3 & 80.5 & 51.4 & 84.1 & 80.7 & 67.8 & 82.5 & 55.1 & 81.2 & 42.5 & \multicolumn{1}{c|}{61.3} & 14.3 & 66.1 & 64.6\\
\multicolumn{1}{l||}{MiB} & \multicolumn{1}{l||}{Transformer} & ViT-small & {\ul 87.8} & 39.2 & {\ul 87.3} & {\ul 61.4} & {\ul 78.3} & 50.0 & 78.0 & 87.7 & {\ul 40.1} & 78.6 & \textbf{58.6} & 73.8 & 86.2 & 72.9 & {\ul 84.4} & {\ul 60.7} & 81.2 & {\ul 53.4} & \multicolumn{1}{c|}{62.9} & {\ul 37.6} & {\ul 74.3} & {\ul 73.8}\\
\multicolumn{1}{l||}{TISS(ours)} & \multicolumn{1}{l||}{Transformer} & ViT-small & \textbf{92.0} & \textbf{45.0} & \textbf{93.8} & \textbf{75.9} & \textbf{85.8} & \textbf{93.1} & \textbf{89.0} & \textbf{93.8} & \textbf{52.9} & \textbf{93.7} & {\ul 57.7} & \textbf{90.0} & \textbf{92.6} & \textbf{86.0} & \textbf{88.5} & \textbf{73.4} & \textbf{92.3} & \textbf{60.2} & \multicolumn{1}{c|}{\textbf{86.5}} & \textbf{50.8} & \textbf{81.8} & \textbf{80.3}\\ \hline
\multicolumn{1}{l||}{offline$\dagger$} & \multicolumn{1}{l||}{Deeplab-v3} & ResNet-101 & 90.2 & 42.2 & 89.5 & 69.1 & 82.3 & 92.5 & 90.0 & 94.2 & 39.2 & 87.6 & 56.4 & 91.2 & 86.8 & 88.0 & 86.8 & 62.3 & 88.4 & 49.5 & \multicolumn{1}{c|}{85.0} & 78.0 & 77.4 & 77.4\\
\multicolumn{1}{l||}{offline} & \multicolumn{1}{l||}{Transformer} & ViT-small & 95.2 & 48.1 & 95.3 & 79.0 & 88.8 & 96.2 & 92.1 & 96.1 & 55.7 & 95.2 & 59.3 & 92.4 & 94.1 & 88.9 & 91.1 & 75.2 & 94.1 & 62.2 & \multicolumn{1}{c|}{88.9} & 79.2 & 82.8 & 81.3\\\hline
\end{tabular}
}

\end{table*}

\begin{table*}[!htbp]

\centering
    \caption{Per Class mIoU on \textbf{19-1} task of Pascal-VOC 2012 under \textsl{overlapped} setup. Best in \textbf{bold}, runner-up {\ul underlined}. $\dagger$: results from \cite{cermelli2020modeling}}
    \label{table:19-1o}
\vskip -0.3cm
\resizebox{\textwidth}{!}{
\begin{tabular}{lll||ccccccccccccccccccc|c||c|c}
\hline
\multicolumn{1}{l||}{\textbf{Method}} & \multicolumn{1}{l||}{\textbf{Architecture}} & \textbf{Backbone} & aero & bike & bird & boat & bottle & bus & car & cat & chair & cow & table & dog & horse & mbike & person & plant & sheep & sofa & \multicolumn{1}{c|}{train} & tv & \textbf{1-19} & \textbf{all}\\ \hline
\multicolumn{1}{l||}{FT$\dagger$} & \multicolumn{1}{l||}{Deeplab-v3} & ResNet-101 & 23.7 & 1.9 & 1.5 & 9.3 & 6.9 & 16.9 & 8.5 & 0.0 & 0.0 & 9.5 & 5.3 & 0.1 & 2.9 & 8.8 & 0.0 & 15.1 & 1.0 & 0.7 & \multicolumn{1}{c|}{16.0} & 12.9 & 6.8 & 7.1\\
\multicolumn{1}{l||}{FT} & \multicolumn{1}{l||}{Transformer} & ViT-small & 20.9 & 1.8 & 1.2 & 9.6 & 5.8 & 15.1 & 9.2 & 0.5 & 0.3 & 6.2 & 5.9 & 0.4 & 0.3 & 6.7 & 0.0 & 14.7 & 0.4 & 0.1 & \multicolumn{1}{c|}{14.9} & 53.1 & 6.0 & 8.4 \\ \hline
\multicolumn{1}{l||}{MiB$\dagger$} & \multicolumn{1}{l||}{Deeplab-v3} & ResNet-101 & 78.1 & 36.2 & 86.8 & 49.4 & 72.7 & {\ul 80.8} & 78.2 & {\ul 90.8} & 38.3 & {\ul 82.0} & 51.9 & {\ul 86.7} & 82.8 & 76.9 & 83.8 & 58.8 & {\ul 84.4} & 45.7 & \multicolumn{1}{c|}{{\ul 68.5}} & 22.1 & 70.2 & 67.8\\
\multicolumn{1}{l||}{ILT$\dagger$} & \multicolumn{1}{l||}{Deeplab-v3} & ResNet-101 & 83.7 & {\ul 40.8} & 80.8 & 59.1 & 58.4 & 77.6 & {\ul 82.4} & 82.3 & 38.9 & 81.7 & 50.8 & 84.8 & {\ul 86.6} & {\ul 81.0} & 83.3 & 56.4 & 82.2 & 43.8 & \multicolumn{1}{c|}{57.5} & 16.4 & 69.1 & 66.4\\
\multicolumn{1}{l||}{MiB} & \multicolumn{1}{l||}{Deeplab-v3} & ResNet-101 & 83.2 & 39.3 & 84.1 & 58.1 & 60.7 & 43.2 & 79.5 & 89.9 & 32.8 & 81.3 & 51.5 & 84.7 & 82.1 & 67.1 & 82.8 & 55.9 & 81.7 & 42.1 & \multicolumn{1}{c|}{61.9} & 13.9 & 66.9 & 65.2\\
\multicolumn{1}{l||}{MiB} & \multicolumn{1}{l||}{Transformer} & ViT-small & {\ul 87.4} & 38.3 & {\ul 87.6} & {\ul 60.7} & {\ul 78.3} & 50.6 & 77.4 & 87.9 & {\ul 40.5} & 77.8 & {\ul 57.9} & 73.8 & {\ul 86.6} & 71.4 & {\ul 84.8} & {\ul 60.9} & 81.1 & {\ul 52.6} & \multicolumn{1}{c|}{62.1} & {\ul 37.2} & {\ul 74.9} & {\ul 73.1}\\
\multicolumn{1}{l||}{TISS(ours)} & \multicolumn{1}{l||}{Transformer} & ViT-small & \textbf{92.1} & \textbf{44.5} & \textbf{93.2} & \textbf{76.1} & \textbf{85.8} & \textbf{93.4} & \textbf{89.3} & \textbf{93.5} & \textbf{50.2} & \textbf{93.7} & \textbf{58.9} & \textbf{89.9} & \textbf{92.4} & \textbf{84.9} & \textbf{88.2} & \textbf{71.3} & \textbf{92.6} & \textbf{59.8} & \multicolumn{1}{c|}{\textbf{85.7}} & \textbf{49.8} & \textbf{81.5} & \textbf{79.9}\\ \hline
\multicolumn{1}{l||}{offline$\dagger$} & \multicolumn{1}{l||}{Deeplab-v3} & ResNet-101 & 90.2 & 42.2 & 89.5 & 69.1 & 82.3 & 92.5 & 90.0 & 94.2 & 39.2 & 87.6 & 56.4 & 91.2 & 86.8 & 88.0 & 86.8 & 62.3 & 88.4 & 49.5 & \multicolumn{1}{c|}{85.0} & 78.0 & 77.4 & 77.4\\
\multicolumn{1}{l||}{offline} & \multicolumn{1}{l||}{Transformer} & ViT-small & 95.2 & 48.1 & 95.3 & 79.0 & 88.8 & 96.2 & 92.1 & 96.1 & 55.7 & 95.2 & 59.3 & 92.4 & 94.1 & 88.9 & 91.1 & 75.2 & 94.1 & 62.2 & \multicolumn{1}{c|}{88.9} & 79.2 & 82.8 & 81.3\\\hline
\end{tabular}
}
\end{table*}

\begin{table*}[!htbp]

\centering
    \caption{Per Class mIoU on \textbf{15-5} task of Pascal-VOC 2012 under \textsl{disjoint} setup. Best in \textbf{bold}, runner-up {\ul underlined}. $\dagger$: results from \cite{cermelli2020modeling}}
    \label{table:15-5d}
\vskip -0.3cm
\resizebox{\textwidth}{!}{
\begin{tabular}{lll||ccccccccccccccc|ccccc||c|c|c}
\hline
\multicolumn{1}{l||}{\textbf{Method}} & \multicolumn{1}{l||}{\textbf{Architecture}} & \textbf{Backbone} & aero & bike & bird & boat & bottle & bus & car & cat & chair & cow & table & dog & horse & mbike & \multicolumn{1}{c|}{person} & plant & sheep & sofa & train & tv & \textbf{1-15} & \textbf{16-20} & \textbf{all}\\ \hline
\multicolumn{1}{l||}{FT$\dagger$} & \multicolumn{1}{l||}{Deeplab-v3} & ResNet-101 & 6.1 & 0.0 & 0.2 & 8.3 & 0.1 & 0.0 & 0.1 & 0.0 & 0.0 & 0.0 & 0.0 & 0.0 & 1.8 & 0.0 & \multicolumn{1}{c|}{0.0} & 24.6 & 24.3 & 36.2 & 32.5 & 50.2 & 1.1 & 33.6 & 9.2\\
\multicolumn{1}{l||}{FT} & \multicolumn{1}{l||}{Transformer} & ViT-small & 45.1 & 0.0 & 46.3 & 28.4 & 14.2 & 40.1 & 41.2 & 0.0 & 0.0 & 34.2 & 0.0 & 0.3 & 13.3 & 4.8 & \multicolumn{1}{c|}{0.0} & 38.7 & 51.8 & 32.6 & 54.5 & 51.5 & 17.9 & 45.8 & 24.9 \\ \hline
\multicolumn{1}{l||}{MiB$\dagger$} & \multicolumn{1}{l||}{Deeplab-v3} & ResNet-101 & 84.4 & 39.4 & 87.5 & 65.2 & 77.8 & 61.0 & 86.0 & 90.9 & {\ul 35.3} & 60.3 & 53.0 & {\ul 88.2} & 80.4 & {\ul 82.4} & \multicolumn{1}{c|}{{\ul 85.3}} & 28.7 & 46.0 & 34.7 & 54.4 & 52.7 & 71.8 & 43.4 & 64.7\\
\multicolumn{1}{l||}{ILT$\dagger$} & \multicolumn{1}{l||}{Deeplab-v3} & ResNet-101 & 79.4 & {\ul 42.0} & 80.5 & 63.9 & {\ul 80.4} & 12.8 & 86.0 & 90.9 & {\ul 35.3} & 60.3 & 53.0 & 83.2 & 73.0 & 80.7 & \multicolumn{1}{c|}{85.0} & 36.9 & 29.9 & {\ul 36.8} & 38.3 & 55.7 & 63.2 & 39.5 & 57.3\\
\multicolumn{1}{l||}{MiB} & \multicolumn{1}{l||}{Deeplab-v3} & ResNet-101 & 77.1 & 40.3 & 83.4 & 60.1 & 70.8 & 24.7 & {\ul 87.9} & {\ul 91.5} & 32.4 & 35.8 & 53.9 & 86.0 & {\ul 81.5} & 81.3 & \multicolumn{1}{c|}{84.8} & 40.7 & 40.7 & 33.1 & 46.2 & {\ul 57.5} & 67.6 & 43.8 & 61.9\\
\multicolumn{1}{l||}{MiB} & \multicolumn{1}{l||}{Transformer} & ViT-small & {\ul 90.4} & 38.2 & {\ul 90.2} & {\ul 73.3} & 79.9 & {\ul 71.2} & 83.4 & 88.7 & {\ul 35.3} & {\ul 69.5} & {\ul 56.1} & 77.7 & 81.3 & 79.9 & \multicolumn{1}{c|}{85.0} & {\ul 52.4} & {\ul 52.1} & 28.7 & {\ul 58.9} & 48.8 & {\ul 76.9} & {\ul 50.7} & {\ul 70.4}\\
\multicolumn{1}{l||}{TISS(ours)} & \multicolumn{1}{l||}{Transformer} & ViT-small & \bf{92.0} & \bf{44.4} & \bf{93.8} & \bf{75.7} & \bf{86.3} & \bf{74.5} & \bf{89.7} & \bf{93.6} & \bf{49.9} & \bf{87.6} & \bf{58.6} & \bf{90.3} & \bf{91.1} & \bf{86.6} & \multicolumn{1}{c|}{\bf{88.0}} & \bf{70.6} & \bf{80.5} & \bf{43.5} & \bf{67.7} & \bf{65.5} & \bf{80.9} & \bf{61.7} & \bf{77.3}\\ \hline
\multicolumn{1}{l||}{offline$\dagger$} & \multicolumn{1}{l||}{Deeplab-v3} & ResNet-101 & 90.2 & 42.2 & 89.5 & 69.1 & 82.3 & 92.5 & 90.0 & 94.2 & 39.2 & 87.6 & 56.4 & 91.2 & 86.8 & 88.0 & \multicolumn{1}{c|}{86.8} & 62.3 & 88.4 & 49.5 & 85.0 & 78.0 & 79.1 & 72.6 & 77.4\\
\multicolumn{1}{l||}{offline} & \multicolumn{1}{l||}{Transformer} & ViT-small & 95.2 & 48.1 & 95.3 & 79.0 & 88.8 & 96.2 & 92.1 & 96.1 & 55.7 & 95.2 & 59.3 & 92.4 & 94.1 & 88.9 & \multicolumn{1}{c|}{91.1} & 75.2 & 94.1 & 62.2 & 88.9 & 79.2 & 84.1 & 74.3 & 81.3\\\hline
\end{tabular}
}

\end{table*}

\begin{table*}[!htbp]

\centering
    \caption{Per Class mIoU on \textbf{15-5} task of Pascal-VOC 2012 under \textsl{overlapped} setup. Best in \textbf{bold}, runner-up {\ul underlined}. $\dagger$: results from \cite{cermelli2020modeling}}
    \label{table:15-5o}
\vskip -0.3cm
\resizebox{\textwidth}{!}{
\begin{tabular}{lll||ccccccccccccccc|ccccc||c|c|c}
\hline
\multicolumn{1}{l||}{\textbf{Method}} & \multicolumn{1}{l||}{\textbf{Architecture}} & \textbf{Backbone} & aero & bike & bird & boat & bottle & bus & car & cat & chair & cow & table & dog & horse & mbike & \multicolumn{1}{c|}{person} & plant & sheep & sofa & train & tv & \textbf{1-15} & \textbf{16-20} & \textbf{all}\\ \hline
\multicolumn{1}{l||}{FT$\dagger$} & \multicolumn{1}{l||}{Deeplab-v3} & ResNet-101 & 13.4 & 0.1 & 0.0 & 15.6 & 0.8 & 0.0 & 0.3 & 0.0 & 0.0 & 0.0 & 0.0 & 0.0 & 0.9 & 0.0 & \multicolumn{1}{c|}{0.0} & 30.9 & 21.6 & 32.8 & 34.9 & 45.1 & 2.1 & 33.1 & 9.8\\
\multicolumn{1}{l||}{FT} & \multicolumn{1}{l||}{Transformer} & ViT-small & 47.1 & 0.0 & 48.2 & 29.3 & 17.2 & 42.3 & 43.0 & 0.0 & 0.0 & 33.9 & 0.0 & 2.3 & 18.9 & 7.8 & \multicolumn{1}{c|}{0.2} & 53.5 & 40.8 & 43.6 & 43.2 & 64.5 & 19.3 & 49.1 & 26.8 \\ \hline
\multicolumn{1}{l||}{MiB$\dagger$} & \multicolumn{1}{l||}{Deeplab-v3} & ResNet-101 & 86.6 & 39.3 & 88.9 & 66.1 & {\ul 80.8} & {\ul 86.6} & {\ul 90.1} & {\ul 92.5} & {\ul 38.0} & 64.6 & 56.4 & {\ul 89.6} & 80.5 & {\ul 86.5} & \multicolumn{1}{c|}{{\ul 85.7}} & 30.2 & 52.9 & 31.3 & {\ul 73.2} & {\ul 59.5} & 75.5 & 49.4 & 69.0\\
\multicolumn{1}{l||}{ILT$\dagger$} & \multicolumn{1}{l||}{Deeplab-v3} & ResNet-101 & 77.4 & 40.3 & 78.9 & 61.9 & 78.7 & 53.5 & 86.1 & 88.7 & 33.8 & 15.9 & 51.1 & 83.2 & 80.2 & 79.8 & \multicolumn{1}{c|}{85.0} & 39.5 & 30.9 & 31.0 & 49.3 & 52.6 & 66.3 & 40.6 & 59.9\\
\multicolumn{1}{l||}{MiB} & \multicolumn{1}{l||}{Deeplab-v3} & ResNet-101 & 88.8 & {\ul 40.7} & 86.5 & 66.0 & 76.0 & 73.8 & 87.1 & 90.7 & 36.6 & 57.3 & 54.7 & 86.3 & {\ul 82.8} & 83.9 & \multicolumn{1}{c|}{85.2} & 34.6 & 42.0 & {\ul 32.2} & 62.4 & 55.6 & 74.2 & 45.3 & 67.3\\
\multicolumn{1}{l||}{MiB} & \multicolumn{1}{l||}{Transformer} & ViT-small & {\ul 91.2} & 38.8 & {\ul 90.9} & {\ul 73.8} & 79.3 & 72.9 & 84.2 & 89.2 & 35.2 & {\ul 69.7} & {\ul 58.2} & 78.6 & 81.5 & 80.7 & \multicolumn{1}{c|}{85.6} & {\ul 53.7} & {\ul 55.7} & 30.6 & 59.3 & 49.2 & {\ul 77.4} & {\ul 53.4} & {\ul 70.9}\\
\multicolumn{1}{l||}{TISS(ours)} & \multicolumn{1}{l||}{Transformer} & ViT-small & \bf{91.3} & \bf{44.4} & \bf{93.4} & \bf{75.8} & \bf{86.0} & \bf{87.0} & \bf{90.6} & \bf{93.6} & \bf{50.1} & \bf{88.3} & \bf{60.3} & \bf{90.8} & \bf{90.7} & \bf{86.9} & \multicolumn{1}{c|}{\bf{88.1}} & \bf{71.2} & \bf{81.9} & \bf{43.3} & \bf{77.8} & \bf{64.9} & \bf{81.9} & \bf{67.8} & \bf{78.5}\\ \hline
\multicolumn{1}{l||}{offline$\dagger$} & \multicolumn{1}{l||}{Deeplab-v3} & ResNet-101 & 90.2 & 42.2 & 89.5 & 69.1 & 82.3 & 92.5 & 90.0 & 94.2 & 39.2 & 87.6 & 56.4 & 91.2 & 86.8 & 88.0 & \multicolumn{1}{c|}{86.8} & 62.3 & 88.4 & 49.5 & 85.0 & 78.0 & 79.1 & 72.6 & 77.4\\
\multicolumn{1}{l||}{offline} & \multicolumn{1}{l||}{Transformer} & ViT-small & 95.2 & 48.1 & 95.3 & 79.0 & 88.8 & 96.2 & 92.1 & 96.1 & 55.7 & 95.2 & 59.3 & 92.4 & 94.1 & 88.9 & \multicolumn{1}{c|}{91.1} & 75.2 & 94.1 & 62.2 & 88.9 & 79.2 & 84.1 & 74.3 & 81.3\\\hline
\end{tabular}
}

\end{table*}

\begin{table*}[!htbp]

\centering
    \caption{Per Class mIoU on \textbf{15-1} task of Pascal-VOC 2012 under \textsl{disjoint} setup. Best in \textbf{bold}, runner-up {\ul underlined}. $\dagger$: results from \cite{cermelli2020modeling}}
    \label{table:15-1d}
\vskip -0.3cm
\resizebox{\textwidth}{!}{
\begin{tabular}{lll||ccccccccccccccc|c|c|c|c|c||c|c|c}
\hline
\multicolumn{1}{l||}{\textbf{Method}} & \multicolumn{1}{l||}{\textbf{Architecture}} & \textbf{Backbone} & aero & bike & bird & boat & bottle & bus & car & cat & chair & cow & table & dog & horse & mbike & \multicolumn{1}{c|}{person} & plant & sheep & sofa & train & tv & \textbf{1-15} & \textbf{16-20} & \textbf{all}\\ \hline
\multicolumn{1}{l||}{FT$\dagger$} & \multicolumn{1}{l||}{Deeplab-v3} & ResNet-101 & 0.3 & 0.0 & 0.0 & 2.5 & 0.0 & 0.0 & 0.0 & 0.0 & 0.0 & 0.0 & 0.0 & 0.0 & 0.9 & 0.0 & \multicolumn{1}{c|}{0.0} & 0.0 & 0.0 & 0.0 & 0.0 & 8.8 & 0.2 & 1.8 & 0.6\\
\multicolumn{1}{l||}{FT} & \multicolumn{1}{l||}{Transformer} & ViT-small & 15.8 & 0.0 & 6.1 & 9.3 & 5.9 & 11.7 & 1.3 & 0.0 & 0.0 & 3.9 & 0.0 & 0.3 & 10.2 & 1.7 & \multicolumn{1}{c|}{1.5} & 29.3 & 42.1 & 26.8 & 38.4 & 42.3 & 4.6 & 35.8 & 12.4 \\ \hline
\multicolumn{1}{l||}{MiB$\dagger$} & \multicolumn{1}{l||}{Deeplab-v3} & ResNet-101 & 53.6 & {\ul 38.9} & 53.6 & 17.7 & 62.7 & {\ul 36.5} & 71.2 & 60.1 & 1.1 & 35.2 & 8.1 & 57.6 & 55.0 & 62.1 & \multicolumn{1}{c|}{79.4} & 10.2 & 14.2 & 11.9 & 18.2 & 10.1 & 46.2 & 12.9 & 37.9\\
\multicolumn{1}{l||}{ILT$\dagger$} & \multicolumn{1}{l||}{Deeplab-v3} & ResNet-101 & 3.7 & 0.0 & 2.9 & 0.0 & 12.8 & 0.0 & 0.0 & 0.1 & 0.0 & 0.0 & 21.2 & 0.1 & 0.4 & 0.6 & \multicolumn{1}{c|}{13.6} & 0.0 & 0.0 & 11.6 & 8.3 & 8.5 & 3.7 & 5.7 & 4.2\\
\multicolumn{1}{l||}{MiB} & \multicolumn{1}{l||}{Deeplab-v3} & ResNet-101 & 35.0 & 30.4 & 42.9 & 24.9 & 57.6 & 12.0 & 46.6 & 80.1 & 0.1 & 25.7 & 14.4 & 67.9 & 62.7 & 48.7 & \multicolumn{1}{c|}{78.6} & 10.4 & 20.2 & 8.9 & 17.1 & 4.2 & 45.6 & 11.3 & 36.6\\
\multicolumn{1}{l||}{MiB} & \multicolumn{1}{l||}{Transformer} & ViT-small & {\ul 69.1} & 36.2 & {\ul 89.1} & {\ul 64.7} & {\ul 77.4} & 9.1 & {\ul 77.7} & {\ul 87.2} & {\ul 31.3} & {\ul 60.6} & {\ul 36.9} & {\ul 68.2} & {\ul 79.0} & {\ul 71.2} & \multicolumn{1}{c|}{{\ul 82.2}} & {\ul 48.1} & {\ul 30.6} & {\ul 24.6} & {\ul 26.3} & {\ul 25.8} & {\ul 68.4} & {\ul 35.8} & {\ul 60.7}\\
\multicolumn{1}{l||}{TISS(ours)} & \multicolumn{1}{l||}{Transformer} & ViT-small & \bf{89.6} & \bf{46.4} & \bf{93.0} & \bf{76.1} & \bf{85.8} & \bf{59.0} & \bf{87.0} & \bf{92.9} & \bf{43.1} & \bf{81.9} & \bf{44.8} & \bf{86.5} & \bf{91.2} & \bf{81.9} & \multicolumn{1}{c|}{\bf{87.3}} & \bf{58.2} & \bf{58.1} & \bf{32.0} & \bf{32.2} & \bf{39.8} & \bf{77.3} & \bf{45.6} & \bf{69.4}\\ \hline
\multicolumn{1}{l||}{offline$\dagger$} & \multicolumn{1}{l||}{Deeplab-v3} & ResNet-101 & 90.2 & 42.2 & 89.5 & 69.1 & 82.3 & 92.5 & 90.0 & 94.2 & 39.2 & 87.6 & 56.4 & 91.2 & 86.8 & 88.0 & \multicolumn{1}{c|}{86.8} & 62.3 & 88.4 & 49.5 & 85.0 & 78.0 & 79.1 & 72.6 & 77.4\\
\multicolumn{1}{l||}{offline} & \multicolumn{1}{l||}{Transformer} & ViT-small & 95.2 & 48.1 & 95.3 & 79.0 & 88.8 & 96.2 & 92.1 & 96.1 & 55.7 & 95.2 & 59.3 & 92.4 & 94.1 & 88.9 & \multicolumn{1}{c|}{91.1} & 75.2 & 94.1 & 62.2 & 88.9 & 79.2 & 84.1 & 74.3 & 81.3\\\hline
\end{tabular}
}

\end{table*}

\begin{table*}[!htbp]

\centering
    \caption{Per Class mIoU on \textbf{15-1} task of Pascal-VOC 2012 under \textsl{overlapped} setup. Best in \textbf{bold}, runner-up {\ul underlined}. $\dagger$: results from \cite{cermelli2020modeling}}
    \label{table:15-1o}
\vskip -0.3cm
\resizebox{\textwidth}{!}{
\begin{tabular}{lll||ccccccccccccccc|c|c|c|c|c||c|c|c}
\hline
\multicolumn{1}{l||}{\textbf{Method}} & \multicolumn{1}{l||}{\textbf{Architecture}} & \textbf{Backbone} & aero & bike & bird & boat & bottle & bus & car & cat & chair & cow & table & dog & horse & mbike & \multicolumn{1}{c|}{person} & plant & sheep & sofa & train & tv & \textbf{1-15} & \textbf{16-20} & \textbf{all}\\ \hline
\multicolumn{1}{l||}{FT$\dagger$} & \multicolumn{1}{l||}{Deeplab-v3} & ResNet-101 & 2.6 & 0.0 & 0.0 & 0.7 & 0.0 & 0.1 & 0.0 & 0.0 & 0.0 & 0.0 & 0.0 & 0.0 & 0.0 & 0.0 & \multicolumn{1}{c|}{0.0} & 0.0 & 0.0 & 0.0 & 0.0 & 9.2 & 0.2 & 1.8 & 0.6\\
\multicolumn{1}{l||}{FT} & \multicolumn{1}{l||}{Transformer} & ViT-small & 15.2 & 0.0 & 7.2 & 9.8 & 5.5 & 12.1 & 2.1 & 0.0 & 0.2 & 4.1 & 0.0 & 0.7 & 10.8 & 1.3 & \multicolumn{1}{c|}{1.9} & 31.4 & 41.8 & 33.8 & 39.6 & 44.7 & 4.7 & 38.3 & 13.1 \\ \hline
\multicolumn{1}{l||}{MiB$\dagger$} & \multicolumn{1}{l||}{Deeplab-v3} & ResNet-101 & 31.3 & 25.4 & 26.7 & 26.9 & 46.1 & 31.0 & 63.6 & 52.8 & 0.1 & 11.0 & 9.4 & 52.4 & 41.2 & 28.1 & \multicolumn{1}{c|}{{\ul 80.7}} & 17.6 & 13.1 & 15.3 & 15.3 & 6.2 & 35.1 & 13.5 & 29.7\\
\multicolumn{1}{l||}{ILT$\dagger$} & \multicolumn{1}{l||}{Deeplab-v3} & ResNet-101 & 20.0 & 0.0 & 3.2 & 6.3 & 2.3 & 0.0 & 0.0 & 0.0 & 0.3 & 5.1 & 19.0 & 0.0 & 9.1 & 0.0 & \multicolumn{1}{c|}{8.7} & 0.0 & 0.0 & 21.0 & 9.9 & 8.1 & 4.9 & 7.8 & 5.7\\
\multicolumn{1}{l||}{MiB} & \multicolumn{1}{l||}{Deeplab-v3} & ResNet-101 & 31.0 & 24.7 & 26.9 & 25.3 & 46.8 & {\ul 31.9} & 62.8 & 52.1 & 1.2 & 10.3 & 9.1 & 49.3 & 42.3 & 27.7 & \multicolumn{1}{c|}{79.8} & 17.9 & 11.2 & 14.9 & 16.2 & 5.8 & 34.0 & 12.7 & 28.9\\
\multicolumn{1}{l||}{MiB} & \multicolumn{1}{l||}{Transformer} & ViT-small & {\ul 68.2} & {\ul 35.5} & {\ul 88.2} & {\ul 65.3} & {\ul 76.5} & 9.9 & {\ul 77.6} & {\ul 84.1} & {\ul 28.7} & {\ul 60.9} & {\ul 35.8} & {\ul 66.7} & {\ul 77.4} & {\ul 70.5} & \multicolumn{1}{c|}{80.1} & {\ul 40.3} & {\ul 31.2} & {\ul 25.8} & {\ul 24.7} & {\ul 25.3} & {\ul 64.9} & {\ul 30.7} & {\ul 60.4}\\
\multicolumn{1}{l||}{TISS(ours)} & \multicolumn{1}{l||}{Transformer} & ViT-small & \bf{89.0} & \bf{46.7} & \bf{92.2} & \bf{76.1} & \bf{85.7} & \bf{80.4} & \bf{88.8} & \bf{93.5} & \bf{44.2} & \bf{86.6} & \bf{36.1} & \bf{89.6} & \bf{88.6} & \bf{84.2} & \multicolumn{1}{c|}{\bf{87.6}} & \bf{53.9} & \bf{77.4} & \bf{38.2} & \bf{41.7} & \bf{44.3} &\bf{ 78.7} & \bf{51.1} & \bf{72.1}\\ \hline
\multicolumn{1}{l||}{offline$\dagger$} & \multicolumn{1}{l||}{Deeplab-v3} & ResNet-101 & 90.2 & 42.2 & 89.5 & 69.1 & 82.3 & 92.5 & 90.0 & 94.2 & 39.2 & 87.6 & 56.4 & 91.2 & 86.8 & 88.0 & \multicolumn{1}{c|}{86.8} & 62.3 & 88.4 & 49.5 & 85.0 & 78.0 & 79.1 & 72.6 & 77.4\\
\multicolumn{1}{l||}{offline} & \multicolumn{1}{l||}{Transformer} & ViT-small & 95.2 & 48.1 & 95.3 & 79.0 & 88.8 & 96.2 & 92.1 & 96.1 & 55.7 & 95.2 & 59.3 & 92.4 & 94.1 & 88.9 & \multicolumn{1}{c|}{91.1} & 75.2 & 94.1 & 62.2 & 88.9 & 79.2 & 84.1 & 74.3 & 81.3\\\hline
\end{tabular}
}

\end{table*}

\subsection{More Results with Larger Backbone}

\begin{table*}[!htbp]

\centering
    \caption{mIoU on the ADE20K dataset for different incremental class learning scenarios. Best in \textbf{bold}, runner-up {\ul underlined}. }
    \label{table:ade-base}
\vskip -0.3cm
\resizebox{\textwidth}{!}{
\begin{tabular}{lllccc||cccccccc||ccccc}
\hline
                                  &            &            & \multicolumn{3}{||c||}{\textbf{100-50}}                                   & \multicolumn{8}{c||}{\textbf{100-10}}                                                                                                                                                       & \multicolumn{5}{c}{\textbf{50-50}}                                                                                                                 \\\hline
\multicolumn{1}{l||}{\textbf{Method}} & \multicolumn{1}{l||}{\textbf{Architecture}} & \textbf{Backbone}   & \multicolumn{1}{||c}{\textsl{1-100}} & \multicolumn{1}{c|}{\textsl{101-150}} & \textsl{all}           & \textsl{1-100} & \textsl{101-110} & \textsl{111-120} & \textsl{121-130} & \textsl{131-140} & \multicolumn{1}{l|}{\textsl{141-150}} & \multicolumn{1}{l|}{\textsl{101-150}} & \textsl{all}  & \multicolumn{1}{l}{\textsl{1-50}} & \textsl{51-100} & \multicolumn{1}{c|}{\textsl{101-150}} & \multicolumn{1}{l|}{\textsl{51-150}} & \textsl{all}  \\ \hline
\multicolumn{1}{l||}{FT}&   \multicolumn{1}{l||}{Transformer}           & ViT-small   & \multicolumn{1}{||c}{1.2}            & \multicolumn{1}{c|}{28.8}             & 10.4          & 0.3            & 0.8              & 0.8              & 1.0              & 1.2              & \multicolumn{1}{c|}{39.7}             & \multicolumn{1}{c|}{8.7}              & 3.1           & 0.5                               & 2.1             & \multicolumn{1}{c|}{35.4}             & \multicolumn{1}{c|}{18.1}            & 12.7           \\
\multicolumn{1}{l||}{FT}&   \multicolumn{1}{l||}{Transformer}           & ViT-base   & \multicolumn{1}{||c}{2.3}            & \multicolumn{1}{c|}{37.7}             & 14.1          & 1.1            & 1.6              & 1.9              & 2.1              & 1.8              & \multicolumn{1}{c|}{43.6}             & \multicolumn{1}{c|}{10.2}              & 4.1           & 2.1                               & 4.8             & \multicolumn{1}{c|}{43.7}             & \multicolumn{1}{c|}{24.3}            & 16.9           \\ \hline
\multicolumn{1}{l||}{MiB}&   \multicolumn{1}{l||}{Transformer}          & ViT-small   & \multicolumn{1}{||c}{39.2}           & \multicolumn{1}{c|}{{28.7}}       & {35.7}    & 35.4           & {19.8}       & {26.4}       & {29.9}       & {14.8}       & \multicolumn{1}{c|}{15.3}             & \multicolumn{1}{c|}{{17.8}}       & 30.7          & 47.8                              & {33.9}      & \multicolumn{1}{c|}{{28.1}}       & \multicolumn{1}{c|}{{30.5}}      & {36.6}          \\
\multicolumn{1}{l||}{MiB}&   \multicolumn{1}{l||}{Transformer}          & ViT-base   & \multicolumn{1}{||c}{\ul 46.1}           & \multicolumn{1}{c|}{{\ul 34.5}}       & {\ul 42.2}    & {\ul 45.8}           & {\ul 22.6}       & {\ul 38.6}       & {\ul 35.5}       & {\ul 19.4}       & \multicolumn{1}{c|}{\ul 24.0}             & \multicolumn{1}{c|}{{\ul 28.0}}       & {\ul 39.9}          & {\ul 51.4}                              & {\ul 38.0}      & \multicolumn{1}{c|}{{\ul 33.3}}       & \multicolumn{1}{c|}{{\ul 35.7}}      & {\ul 40.9}          \\
\multicolumn{1}{l||}{TISS(ours)}&   \multicolumn{1}{l||}{Transformer}   & ViT-base   & \multicolumn{1}{||c}{\textbf{47.2}}  & \multicolumn{1}{c|}{\textbf{36.2}}    & \textbf{43.5} & \textbf{46.5}  & \textbf{23.8}    & \textbf{39.1}    & \textbf{38.6}    & \textbf{25.6}    & \multicolumn{1}{c|}{\textbf{25.1}}    & \multicolumn{1}{c|}{\textbf{30.4}}    & \textbf{41.1} & \textbf{52.7}                     & \textbf{39.3}   & \multicolumn{1}{c|}{\textbf{34.2}}    & \multicolumn{1}{c|}{\textbf{36.8}}   & \textbf{42.1} \\ \hline
\multicolumn{1}{l||}{offline}&   \multicolumn{1}{l||}{Transformer}      & ViT-small   & \multicolumn{1}{||c}{45.2}           & \multicolumn{1}{c|}{29.1}             & 39.9          & 45.2           & 27.2             & 43.1             & 32.4             & 29.3             & \multicolumn{1}{c|}{17.1}             & \multicolumn{1}{c|}{29.1}             & 39.9          & 52.8                              & 39.1            & \multicolumn{1}{c|}{28.6}             & \multicolumn{1}{c|}{32.1}            & 39.9         \\
\multicolumn{1}{l||}{offline}&   \multicolumn{1}{l||}{Transformer}      & ViT-base   & \multicolumn{1}{||c}{52.7}           & \multicolumn{1}{c|}{35.0}             & 46.8          & 52.7           & 30.3             & 48.4             & 35.6             & 32.3             & \multicolumn{1}{c|}{28.4}             & \multicolumn{1}{c|}{35.0}             & 46.8         & 55.8                             & 49.6            & \multicolumn{1}{c|}{35.0}             & \multicolumn{1}{c|}{42.3}            & 46.8         \\
\hline
\end{tabular}
}

\end{table*}

We build \textsl{MiB}\cite{cermelli2020modeling} and \textsl{TISS} upon larger backbone, Vision Transformer(ViT)\cite{dosovitskiy2020image} Base(denoted as ViT-base) model with 86 million parameters. As Table \ref{table:ade-base} shows, for \textbf{100-50} task, \textsl{TISS} surpasses \textsl{MiB} by $2.4\%$ in the first 100 classes, $4.9\%$ in the added 50 classes and $2.6\%$ in all classes. When it comes to \textbf{100-10} task, \textsl{TISS} achieves an average improvement over \textsl{MiB} of $1.5\%$ in the first 100 classes, $6.2\%$ in the added 50 classes and $2.5\%$ in all classes. Finally, in \textbf{50-50} task, \textsl{TISS} surpasses over \textsl{MiB} of $3.5\%$ in the first 50 classes, $6.9\%$ in the added 100 classes and $4.2\%$ in all classes.

\end{document}